\documentclass[lettersize,journal]{IEEEtran}
\usepackage{amsmath,amsfonts,amssymb}
\usepackage{mathtools}
\usepackage{algorithmic}
\usepackage{algorithm}
\usepackage{array}
\usepackage[caption=false,font=normalsize,labelfont=sf,textfont=sf]{subfig}
\usepackage{textcomp}
\usepackage{stfloats}
\usepackage{url}
\usepackage{verbatim}
\usepackage{graphicx}
\usepackage{cite}
\usepackage{xcolor}
\usepackage{multirow}
\usepackage{diagbox}
\usepackage{hhline}
\usepackage{tikz}
\usepackage[backref=page]{hyperref}
\hypersetup{
	colorlinks=true,
	linkcolor=[rgb]{0.1796862745, 0.3134117647, 0.9127843137},
	citecolor=green,
	filecolor=magenta,      
	urlcolor=gray,
}
\graphicspath{{figures/}}

\usepackage{microtype}
\DisableLigatures[f]{encoding = *, family = *}

\begin{document}

\newcommand{\network}[1][s]{PadéNet#1}
\newcommand{\layer}[1][s]{$\operatorname{\textit{PaLa}}$#1}
\newcommand{\neuron}[1][s]{$\operatorname{\textit{Paon}}$#1}
\newcommand{\relu}{$\operatorname{ReLU}$}
\newcommand{\mainact}{$\operatorname{GELU}$}
\newcommand{\pau}{$\operatorname{PAU}$}
\newcommand{\f}[1]{\textcolor{red}{\underline{#1}}}
\newcommand{\s}[1]{\textcolor{blue}{\underline{#1}}}
\newcommand{\shifter}{$\operatorname{Shifter}$ }

\title{Padé Neurons for Efficient Neural Models}

\author{Onur Keleş,~\IEEEmembership{Member,~IEEE,} and A. Murat Tekalp,~\IEEEmembership{Life Fellow,~IEEE}
\thanks{O.K. is with the Department of Electrical and Electronics Engineering, Koç University, İstanbul, Türkiye, and Codeway AI Research. \\
\indent A.M.T. is with the~Department of Electrical and Electronics Engineering, Koç University, İstanbul, Türkiye.}
\thanks{A.M.T. acknowledges support from Turkish Academy of Sciences (TUBA).}
\thanks{This paper is a  significantly expanded version of our ICIP 2024 paper~\cite{keles2024paon}. This manuscript presents superior image super-resolution results due to~improvements in the Paon model and completely new image compression and classification results.} 
}

\markboth{Accepted for Publication in IEEE TRANSACTIONS ON IMAGE PROCESSING}%
{Shell \MakeLowercase{\textit{et al.}}: A Sample Article Using IEEEtran.cls for IEEE Journals}


\maketitle

\begin{abstract}
Neural networks commonly employ the~McCulloch-Pitts neuron model, which is a linear model followed by a point-wise non-linear activation. Various researchers have already advanced inherently non-linear neuron models, such as quadratic neurons, generalized operational neurons, generative neurons, and super neurons, which offer stronger non-linearity compared to point-wise activation functions. In this paper, we introduce a novel and better non-linear neuron model called Padé neurons (\neuron[s]), inspired by Padé approximants. \neuron[s] offer several advantages, such as diversity of non-linearity, since each \neuron[] learns a different non-linear function of its inputs, and layer efficiency, since \neuron[s] provide stronger non-linearity in much fewer layers compared to piecewise linear approximation. Furthermore, \neuron[s] include all previously proposed neuron models as special cases, thus any neuron model in any network can be replaced by \neuron[s]. We note that there has been a proposal to employ the Padé approximation as a generalized point-wise activation function, which is fundamentally different from our model. To validate the~efficacy of \neuron[s], in our experiments, we replace classic neurons in some well-known neural image super-resolution,  compression, and classification models based on the ResNet architecture with \neuron[s]. Our comprehensive experimental results and analyses demonstrate that neural models built by \neuron[s] provide better or equal performance than their classic counterparts with a smaller number of layers. The PyTorch implementation code for \neuron[] is open-sourced at \href{https://github.com/onur-keles/Paon}{\url{https://github.com/onur-keles/Paon}}.
\end{abstract}


\section{Introduction}
\label{sec:intro}

\IEEEPARstart{D}{espite} the popularity of deep neural networks surged only about a decade ago, the foundational concepts underlying their fundamental unit, the neuron model, have been established for quite some time. The classical McCulloch-Pitts neuron \cite{mcculloch1943logical, rosenblatt1957perceptron} operates by linearly combining each input element with specific weights and subsequently applying a non-linear activation function to the result. Subsequent research for more powerful neuron models have focused on proposing better behaving point-wise activation functions while keeping the linear component of the model unchanged. The rectified linear unit (\relu)~\cite{krizhevsky2012imagenet} and its variants such as the parametric \relu~\cite{he2015delving} and the Gaussian error linear unit (\mainact) \cite{hendrycks2016gaussian} are among the most widely adopted activation functions. Recognizing that these non-linearities are manually predefined and fixed for all neurons, the study~\cite{molina2019pade} proposed learning a different point-wise activation function for each layer through Padé approximants, initializing the coefficients of the Padé approximation from that of a pre-selected non-linearity.

An alternative line of research has advocated that a neuron model should not be limited only to a point-wise non-linearity through the activation function, and proposed inherently non-linear neuron models. These include: quadratic neurons \cite{cheung1991rotational, milenkovic1996annealing, xu2022quadralib, chen2023expressivity}, which operate on the first and second powers of their inputs; generalized operational perceptrons~\cite{kiranyaz2020operational}, which replace the weighted linear combination and addition operations in the classical neuron with a variety of mathematical functions; generative neurons~\cite{kiranyaz2021self}, which employ Taylor series expansion for polynomial approximation of arbitrary non-linear functions and operate on higher-order powers of the input. Super neurons~\cite{kiranyaz2023super} aim to enhance the receptive field of generative neurons by applying learnable shifts to convolution kernels. The recently proposed Kolmogorov-Arnold networks (KAN)~\cite{liu2024kan} also employ non-linear functions of inputs.

This paper introduces a novel and more powerful, inherently non-linear neuron model called \neuron[], based on Padé approximants. It is well-known that the Padé approximation, which represents arbitrary functions as ratio of two polynomials, often yields a better approximation than truncating the Taylor series expansion of the function, and may still work where the Taylor series do not converge. Hence, \neuron[] is a robust non-linear neuron model without the need for \emph{a fixed external activation function}. Equipped with two variants of \shifter\ module, \neuron[s] can benefit from an expanded receptive field, which improves the performance of convolutional models. Our extensive experiments on single image super-resolution, image compressionand image classification tasks demonstrate~fewer layers of \neuron[s] do provide better performance than other neuron models. 

Our contributions in this work can be summarized as: 
\begin{itemize}
	\item We introduce a novel inherently non-linear neuron model, \neuron[], inspired by Padé approximants. 
    \item We show that  \neuron[] is a super set of previously known neuron models; hence, it can replace any neuron model in any neural network.
	\item We propose two variants of a \shifter module, which are improved versions of the one presented in \cite{keles2024paon} to increase the receptive field of \neuron[s] when used in convolutional neural networks
	\item Our extensive experiments on image super-resolution, image compression and image classification tasks demonstrate that neural models built by \neuron[s] \emph{without using a fixed activation function} provide better or equal performance with fewer layers compared to models based on classical neurons.
    \item We provide results to show that \neuron[s] are resilient to lower precision implementations, which makes them suitable for possible real-world deployment across different platforms.

\end{itemize}

\section{Related Work}
\label{sec:related_work}

There have been numerous efforts to develop more robust activation functions, as well as inherently non-linear neuron models. These attempts aim to enhance the representation capabilities of neural networks. The following sections briefly inspect these activation functions and neuron models.

\subsection{Quadratic Neurons}
\label{subsec:quadtratic_neuron}

The output of a quadratic neuron depends on both the input $x$ and its square $x^2$ given by
\begin{equation}
	\label{eq:general_quadratic_neuron}
	f(x) = A(x^{2}) + B(x),
\end{equation}
where $A$ is a function of $x^2$, and $B$ is a linear function of $x$. The bias term is intentionally excluded to maintain simplicity. This approach allows the neuron to capture more complex patterns in the data compared to traditional linear neurons.

This formulation has been utilized in several studies. Cheung and Leung \cite{cheung1991rotational} employed $f(x)=x^{\text{T}}w_{1}x + w_{2}x$. The~authors of \cite{milenkovic1996annealing} adjusted the second term to $w_{2}x^{2}$. In~\cite{bu2021quadratic}, a quadratic expression was derived by multiplying two filtered inputs, represented as $(w_{1}x)\odot(w_{2}x)$, where $\odot$ denotes element-wise (Hadamard) product. The study \cite{xu2022quadralib} extended this expression by adding $w_{3}x$. Additionally, \cite{chen2023expressivity} applied a low-rank approximation to compute the quadratic terms.

In contrast to the quadratic neuron, the proposed \neuron[s] are not limited to only second-order polynomials, thereby potentially capturing more complex relationships in the data.

\subsection{Generalized Operational Perceptrons}
\label{subsec:gops}

Kıranyaz et al. \cite{kiranyaz2020operational} introduced the generalized operational perceptron model, where linear scaling of inputs with weights and subsequent addition of these results in classic neurons are replaced by nodal and pooling operators, respectively.  The~\textquotedblleft nodal\textquotedblright\ operators, include exponentiation, sinusoidal functions, etc. in addition to the traditional linear scaling by weights. The \textquotedblleft pool\textquotedblright\ operation, which is addition in regular neurons, may be replaced by other operations, such as a median operator. However, choosing and applying these complex operations require significantly more resources than traditional addition and multiplication. Additionally, the selection of these operations is highly dependent on the specific architecture. For instance, if another layer is added to the network, a new search must be conducted to determine the appropriate operations for the new configuration.

\subsection{Generative Neurons}
\label{subsec:generative_neuron}

Recognizing the substantial computational demand of generalized operational perceptrons, Kıranyaz et al. \cite{kiranyaz2021self} introduced generative neurons. These neurons aim to approximate the required mapping function using a truncated Taylor series expansion around the point $0$, essentially applying a Maclaurin series expansion up to a predetermined order. This approach seeks to mitigate the computational burden while still capturing a similar level of non-linearity. However, generative neurons face certain limitations. Since they are linear combinations of different positive orders of the input, their outputs can exceed safe computational ranges. Moreover, Taylor series approximation becomes less accurate as we move away from the point of expansion. To address these issues, the outputs of generative neurons \cite{kiranyaz2021self} are constrained by a $\tanh$ activation function, which is known to cause vanishing gradients and hinder the training of deep models. Even with these limitations, it has been shown that generative neurons provide performance improvements over classical neurons in image super-resolution and compression tasks \cite{keles2021self, yilmaz2021self}.

In contrast, the proposed \neuron[s] calculate higher-order approximations as a ratio of two polynomials. This feature often eliminates the need for limiting activation functions, allowing \network[s] to utilize common non-linearities that effectively overcome the vanishing gradient problem. Moreover, inherent non-linearity that \neuron[s] have might even eliminate the need for an external activation. Additionally, for a given approximation order, the Padé approximant more closely follows a target transcendental function compared to a Taylor series expansion around a point \cite{baker1996pade}. Consequently, \neuron[s] provide a more efficient means of achieving the same level of non-linearity.

\subsection{Enlarging the Receptive Field in Convolutional Networks}
\label{subsec:super_neurons}

A well-known problem with convolutional neural networks is that each neuron has a limited receptive field. Deformable convolutions were proposed~\cite{dai2017deformable} to address this problem in the case of classic neuron models. To enhance the receptive field of generative neurons, super neurons \cite{kiranyaz2023super} were proposed. Super neurons introduce shifts, which are randomly initialized and then optimized through back propagation during training.

In contrast, we introduce an improved \shifter module, which allows \neuron[s] to learn the shifts more effectively from the data, potentially leading to better performance.

\subsection{Padé Activation Unit (PAU)}
\label{subsec:pau}

Molina et al. \cite{molina2019pade} proposes to use the Padé approximant as an activation function, termed the Padé activation unit (\pau). In this approach, the orders of the rational polynomials and some initial coefficients for the desired activation function are pre-determined to provide an initial non-linearity. However, in \pau, the activation function is learned for an entire layer and tends to retain the general shape of the non-linearity whose Padé approximant was used as the starting point for the coefficients.

In contrast, \network[s] adopt a finer approach, where each neuron learns its own rational approximation. This method offers a higher degree of freedom in choosing non-linearities and provides element-wise non-linearity. This capability enhances the flexibility and expressiveness of the neural network.

\section{Padé Approximant Neurons (Paons)}
\label{sec:paon}

In this section, we propose the Padé neuron and analyze its key features. First, we introduce the mathematical formulation of the Padé approximant neuron in Section~\ref{subsec:paon_formulation}. Then, in Section~\ref{subsec:pade_singularity}, we investigate potential singularities and propose solutions to mitigate these issues for stable and reliable computations. In Section~\ref{subsec:shifter}, we  introduce the~\shifter module, which enhances the receptive field of \neuron[]. The~computational complexity of \neuron[] is discussed in Section~\ref{subsec:paon_hyper}. Finally, in Section~\ref{subsec:paon_superset}, we show how \neuron[] encompasses and generalizes the capabilities of previous neuron models.

\subsection{Mathematical Formulation}
\label{subsec:paon_formulation}

In the univariate case, the Padé approximant $f_{[K/L]}(x)$ of a function~$f(x)$ is given by the ratio of two polynomials:
\begin{equation}
	\label{eq:pade_approx}
	f_{[K/L]}(x) = \dfrac{\displaystyle\sum_{k=0}^{K}a_kx^k}{ 1+\displaystyle\sum_{k=1}^{L}b_kx^k} =  \dfrac{a_{0} + \displaystyle\sum_{k=1}^{K}a_kx^k}{1 + \displaystyle\sum_{k=1}^{L}b_kx^k} %
\end{equation}
where $K$ and $L$ are the orders and $a_k$s and $b_k$s are the~coefficients of the numerator and denominator polynomials, respectively, such that $b_0 = 1$. It is the \textquotedblleft best\textquotedblright\ approximation of $f(x)$ by a rational function of given order as shown by the Montessus de Ballore theorem, which establishes uniform convergence of Padé approximants on compact subsets excluding the poles~\cite{mbtheorem1988}. In particular, Eq.~\ref{eq:pade_approx} provides a good approximation of $f(x)$ outside the disk of convergence of the~Taylor series expansion of $f(x)$.

There has been several works to extend this result to multivariate functions~\cite{mvpa1999,mvpa2000}, which address several challenges for proofs of convergence under certain assumptions. Since neurons are functions of several variables (pixels within their receptive fields), we define a Padé neuron of order $[K/L]$ as:
\begin{equation}	\label{eq:vanilla_pade_neuron}
    \operatorname{\textit{Paon}}_{[K/L]}(x(n_1,n_2)) = \dfrac{P_{K}(n_1,n_2)}{Q_{L}(n_1,n_2)}  
\end{equation}
where 
\begin{equation}
    P_{K}(n_1,n_2)= a_{0} + \displaystyle\sum_{k=1}^{K} a_{k}(n_1,n_2) \circledast (x(n_1,n_2))^k
\nonumber
\end{equation}
and
\begin{equation}
    Q_{L}(n_1,n_2)= 1 + \displaystyle\sum_{k=1}^{L} b_{k}(n_1,n_2) \circledast (x(n_1,n_2))^k
\nonumber
\end{equation}
in which $x(n_1,n_2)$ denotes the input of the \neuron[], $a_{k}$ and $b_{k}$ are the weights for the $k$th power of the input in $P_{K}(n_1,n_2)$ and $Q_{L}(n_1,n_2)$, respectively, and $a_0$ is the bias term. 

We refer to network layers consisting of \neuron[s] as Padé Layers (\layer[]). \layer[s] can be fully-connected, where the~operation $\circledast$ in Eq.~\ref{eq:vanilla_pade_neuron} is multiplication, or convolutional, where $\circledast$ is convolution and the parameters $a_{k}$ and $b_{k}$ are shared for all $(n_1,n_2)$. The~implementation of a \layer[] is illustrated for $[K/L] = [2/3]$ in Fig.~\ref{fig:pade_neuron}.

\subsection{Smoothed Padé Approximants to Avoid Singularity}
\label{subsec:pade_singularity}

One critical aspect of the Padé approximants is the potential for the denominator $Q_{L}(n_1,n_2)$ to become equal to or very close to zero. While proper weight initialization can prevent this issue in the beginning, gradient descent learning does not guarantee that the denominator will remain nonzero throughout training. Hence, we propose a smoothed \neuron[], called \neuron[]$^S_{[K/L]}$, inspired by the work of Beckermann and Kalyagin \cite{beckermann1997diagonal}, in order to ensure that the divisor is always nonzero. It is given by:
\begin{equation}
\label{eq:pade_neuron_smooth}
	\operatorname{\textit{Paon}}^S_{[K/L]} =\dfrac{Q_{L}P_{K} + Q_{L-1}P_{K-1}}{Q_{L}^{2} + Q_{L-1}^{2}}
\end{equation}
where $P_{K}$ and $Q_{L}$ are defined as in Eq.~\eqref{eq:vanilla_pade_neuron}, $Q_{L-1}$ and $P_{K-1}$ denote polynomials of one degree lower than $Q_{L}$ and $P_{K}$, respectively, and the indices $(n_1,n_2)$ are not shown for concise notation. Note that, considering Eq.~\eqref{eq:vanilla_pade_neuron}, when the polynomials are of degree zero, the numerator simplifies to the bias term, $P_{0}(x) = a_{0}$, and the denominator becomes $Q_{0}(x) = 1$. Although this method was proposed for Padé approximants with $K=L$, we have observed that it can effectively be applied in cases where $\lvert K-L\rvert=\{0,1\}$ without any modification. This variant ensures smoother behavior and stability by preventing the denominator from approaching zero, maintaining robust computational properties throughout the training process. 

The smoothed \neuron[], \neuron[]$^S_{[K/L]}$, inherently possesses stronger non-linearity than \neuron[]$_{[K/L]}$. A closer examination of Eq.  \eqref{eq:vanilla_pade_neuron} reveals that an $[K/L]$ Padé approximant agrees with a Taylor series of order $K+L$, and requires $K+L$ convolutions in total. However, the~approximation in Eq.~\eqref{eq:pade_neuron_smooth} corresponds to an $[(K+L)/2L]$ Padé approximant, which agrees with a Taylor series of order $K+3L$. Remarkably, this higher-order approximation is achieved with only $K+L$ convolutions. Furthermore, \neuron[]$^S_{[K/L]}$ does not require the~outputs to be bounded by fixed  activation functions, such as $\tanh$, thus, avoiding the~vanishing gradient problem.

\subsection{Shifter Module}
\label{subsec:shifter}

\begin{figure*}[th]
	\centering
	\includegraphics[width=0.9\textwidth]{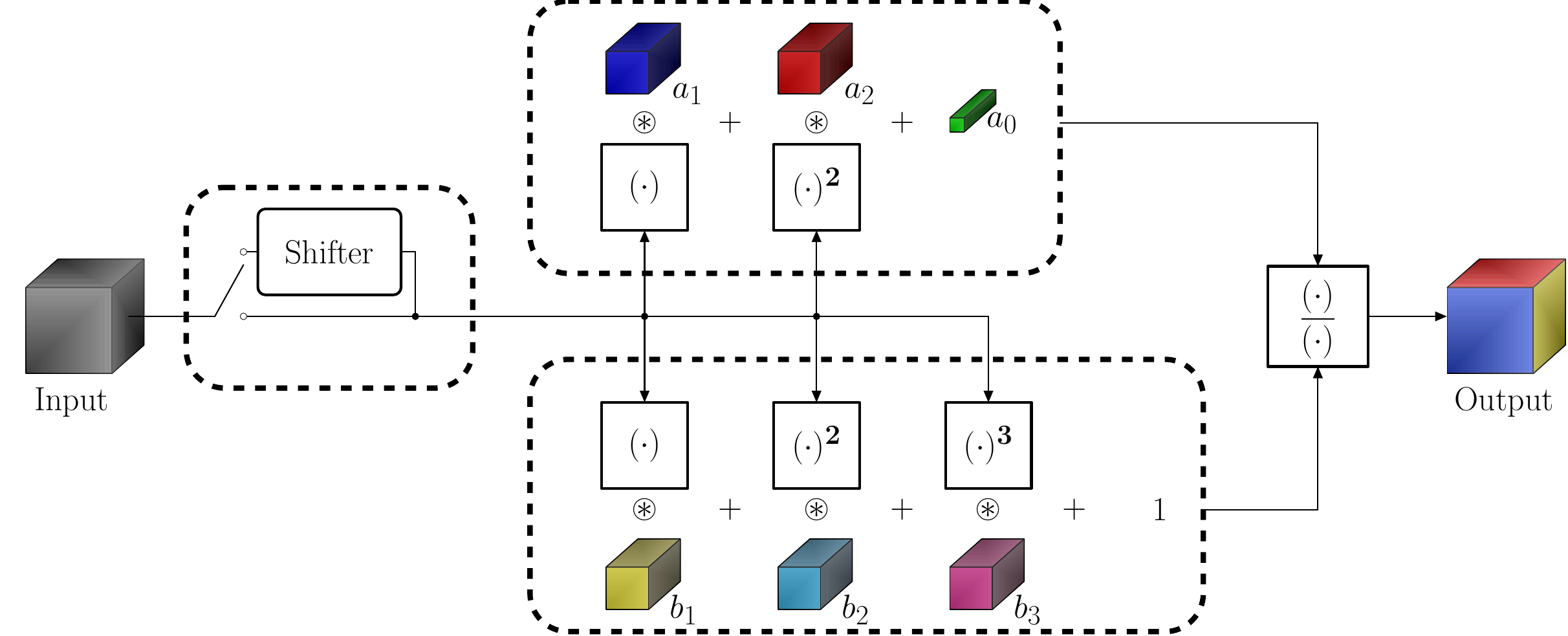}
	\caption{Illustration of a Padé neuron (\neuron[]) for $[K/L] = [2/3]$, where $a_{0}$ is bias for numerator, $ (\cdot)^{k} $ takes $k^{\text{th}}$ power of the~input in element-wise manner, $\frac{(\cdot)}{(\cdot)}$ implements Eq. \eqref{eq:pade_neuron_smooth}. The \shifter module shifts the input features when $\circledast$ is convolution.}
	\label{fig:pade_neuron}
\end{figure*}

The aim of the \shifter is to shift the input features in the~horizontal and/or vertical directions to increase the~receptive field of \neuron[]$^S_{[K/L]}$ such that the convolutions can extract more representative features compared to the case when the input is not shifted. The block diagram of a \neuron[] with a \shifter for $[K/L] = [2/3]$ is illustrated in Fig. \ref{fig:pade_neuron}. We propose two methods for \shifter\hspace{-3pt} in convolutional \layer[], where the~features are shifted kernel-wise (as a group) or in element-wise manner. 

In the first method (also presented in \cite{keles2024paon}), the operation of the \shifter depends on the shift parameter~$b$:
(i) When~$b < 0$, the module is deactivated.
(ii) When $b$ is a positive integer, the module performs gradient-based search to find the optimal shift within the range $[-b, b]$, just as in the super neuron.
(iii)~When $b=0$, the module computes the best shift for each channel without any restrictions. It consists of an averaging operation, a $1\times1$ convolution, and a non-linear activation function, with some constraints to maintain shape consistency.

In the second method, input features are shifted by using deformable kernels~\cite{dai2017deformable}. This approach is more powerful because it can calculate input feature shifts for each weight individually. Deformable convolution allows adjusting the receptive field adaptively for each spatial location. It is crucial to limit the magnitude of shifts to ensure stability by preventing deformable convolution kernel to operate on regions outside of the input feature map. When the shift parameter $b$ is not positive, the limit for the maximum allowable shift $m$ is $\max(h, w)/4$, where $h$ and $w$ are the height and width of the input feature map, respectively. For a positive integer $b$, $m=b$. The limitation of the shift in deformable convolution is achieved using the $\tanh$ operation in the form of $m\cdot\tanh(x/m)$, where $x$ is the offset map. The scaling of $x$ by $m$ is crucial, though often overlooked in the literature \cite{chan2022basicvsr++}, as it prevents small values from being pulled towards the upper limit. Without this scaling, $\tanh$ function may start to saturate even for relatively small input values, leading to unintended shifts. Thus, this method ensures that the shifts remain within a controlled range, maintaining stable and effective training. To ensure that the module only learns shifts that improve the performance, the convolution weights and bias in the \shifter module are initialized to zero. This initialization allows the module to start with a neutral state, learning the necessary shifts only when they improve the model's performance.

For both shifting methods, proper handling of locations outside the original feature map is crucial. The common default approach for handling the boundaries is to pad the~image/feature map with zeros. However, this leads to inaccurate results around the boundaries. To address this issue, during both shifts and convolutions, we pad the input using pixel replication at the borders, as suggested by \cite{tekalp1985boundary}, which ensures that the kernel always operates on meaningful data. 
\vspace{-8pt}

\subsection{Computational Complexity of Paon\texorpdfstring{$^S_{[K/L]}$}{ S[K/L]}}
\label{subsec:paon_hyper}

We calculate the number of multiplications and divisions in the convolutional \layer[] setting. We only consider the~complexity of forward propagation since back-propagation is typically implemented automatically and efficiently by the~PyTorch \cite{paszke2019pytorch} framework. The analysis is performed assuming the input shape is $W\times H\times C_i$, where $W$ and $H$ are the~width and height of the image, and $C_i$ is the number of input channels. 

Let $C_o$ denote the number of \neuron[s] in a $PaLa$, all with $k\times k$~kernels, where $k$ is odd, for convolutions with stride~$1$. If~the input is padded so that its height and width  become $(H+2\cdot\lfloor k/2\rfloor)$ and $(W+2\cdot\lfloor k/2\rfloor)$, respectively, where $\lfloor\cdot\rfloor$ denotes the floor operation, the output shape becomes $W\times H\times C_o$. In this case, a \neuron[]$^S_{[K/L]}$ performs $(K+L)\times C_i\times k\times k\times W\times H\times C_o$ multiplications for convolutions and $W\times H\times C_o$ divisions for calculating the ratio. In addition, we need $4\times W\times H\times C_o$ more operations for tensor multiplications to calculate $Q_{L}(n_1,n_2), P_{K}(n_1,n_2)$, $Q_{L-1}(n_1,n_2), P_{K-1}(n_1,n_2)$, $Q_{L}^2(n_1,n_2)$ and $Q_{L-1}^2(n_1,n_2)$.

For the \shifter \hspace{-4pt}, a convolutional layer calculates the amount of shift in the horizontal and vertical directions for each channel. This requires an additional $2C_i\times k_s\times k_s\times C_i$ parameters, where $k_s$ is the kernel size for the \shifter\hspace{-3pt}. To obtain the shifts as a tensor with dimensions $W\times H\times 2C_i$, the input is padded to the size $(W+2\cdot\lfloor k_s/2\rfloor)\times (H+2\cdot\lfloor k_s/2\rfloor)$. Then, the total number of multiplications required to perform convolution is $2C_i\times k_s\times k_s\times W\times H\times C_i$. The dimensions of the output of \shifter remain the same as the input; i.e., $W\times H\times C_i$. Consequently, $4\times W\times H\times C_i$ more operations are needed for the \shifter to perform bilinear interpolation by calculating weighted sum of $4$ feature \textquotedblleft pixels\textquotedblright\ for each feature element.

Complexity analysis suggests that the number of operations required for \neuron[]$^S_{[K/L]}$ is nearly $(K+L)$ times the number of operations required for the classic neuron. This is verified by a numerical example in the following. For this purpose, two different methods, namely \texttt{fvcore}\footnote{\url{https://github.com/facebookresearch/fvcore}}, which gives the number of MACs, and a native PyTorch method \texttt{torch.utils.flop\_counter}, giving the number of FLOPs, are used. In the example, the input and output are tensors with dimensions $1\times3\times256\times256$, the kernel size is $5\times5$, and the Padé approximation degrees are $[1/1]$. The~notation \neuron[]$^S_{[1/1]}$-I indicates that the first \shifter is used with the parameter $b$. The results are presented in Table \ref{tbl:operational_complexity}\footnote{The number of MACs, FLOPs and peak memory allocation for the deformable convolution and \neuron[-S] with the second \shifter type are not reported since we believe the PyTorch method used to calculate them does not provide reliable results for deformable convolution.}.

\begin{table}
	\centering
	\caption{Number of MACs (given by \texttt{fvcore}) and FLOPS (given by \texttt{flop\_counter}) in millions, and the peak memory allocation (given by \texttt{max\_memory}) in mebibytes (MiB).}
	\resizebox{0.48\textwidth}{!}{
		\begin{tabular}{|c||c|c|c|} 
			\hline
			\diagbox{Layer}{Method} & \texttt{fvcore} & \texttt{flop\_counter} & \texttt{max\_memory}\\\hhline{|=||=|=|=|}
			Classical Neuron & $14.74$ & $29.49$ & $449.59$ \\\hline
			\neuron[]$^S_{[1/1]}$-I($b<0$) & $29.49$ & $58.98$ & $457.89$ \\\hline
			\neuron[]$^S_{[1/1]}$-I($b=0$) & $30.28$ & $58.98$ & $576.85$ \\\hline
			\neuron[]$^S_{[1/1]}$-I($b>0$) & $30.28$ & $58.98$ & $576.85$ \\\hline
		\end{tabular}
	}
	\label{tbl:operational_complexity}
\end{table}

We observe from Table \ref{tbl:operational_complexity} that the number of FLOPs are nearly twice the number of MACs as expected,, since one MAC is equal to one multiplication plus one accumulation,  and that the number of operations required for \neuron[]$^S_{[K/L]}$ is nearly $(K+L)$ times the number of operations required for the classical neurons. 

We also conduct a memory footprint analysis by using the native PyTorch method \texttt{torch.cuda.max\_memory\_allocated} during forward operations, under the same settings used for FLOP and MAC calculations. As expected, the numbers given in Table \ref{tbl:operational_complexity} show that \neuron[]$^S_{[1/1]}$-I requires more memory than its classical counterpart. However, the difference between the memory usage of the classical convolutional neuron and \layer[] without any shift is not proportional to $(K+L)$, as is the case for the number of required operations, but remains relatively small. This is due to PyTorch's efficient tensor management and the relatively small number of additional activations stored for \neuron[] under these settings. In contrast, once the \shifter is activated, the memory increase becomes more pronounced due to the additional forward pass required to calculate the shift values. Naturally, these values may vary depending on settings such as number of input or output channels.

\subsection{Paons as a Super Set of Other Neuron Models}
\label{subsec:paon_superset}

\neuron[] is a super set of the following neuron models, offering adaptability in various configurations:
\begin{itemize}
	\item \textbf{Ordinary Neuron:} For $K=1$, $L=0$, with the \shifter deactivated, the \neuron[] reduces to an ordinary neuron. Note~that $Q_{L}(x)=1$ for $L=0$.
	\item \textbf{Quadratic Neuron:} When $K=2$, $L=0$, the \neuron[] exhibits the properties of a quadratic neuron.
	\item \textbf{Generative Neuron:} For $K\geq2$ and $L=0$, the \neuron[] behaves as a generative neuron, approximating functions via Taylor series expansion using higher-order terms.
	\item \textbf{Super Neuron:} When the \shifter module is activated in the generative neuron setting,  \neuron[] operates as an improved version of the super neuron, learning effective shifts from the input data.
\end{itemize}
Given these properties, \neuron[s] can seamlessly replace any neuron model in any neural network.

\section{Experiments}
\label{sec:exp}

We provide experimental results to show that replacing classic convolution layers with \layer[] (consisting of \neuron[s]) in well-known SISR, image compression, and image classifier models with the~ResNet architecture provides improved performance with a smaller number of layers. We present SISR results that are better than in \cite{keles2024paon} and new image compression and classification results.

\begin{figure*}[ht]
	\centering
	\includegraphics[width=0.85\textwidth]{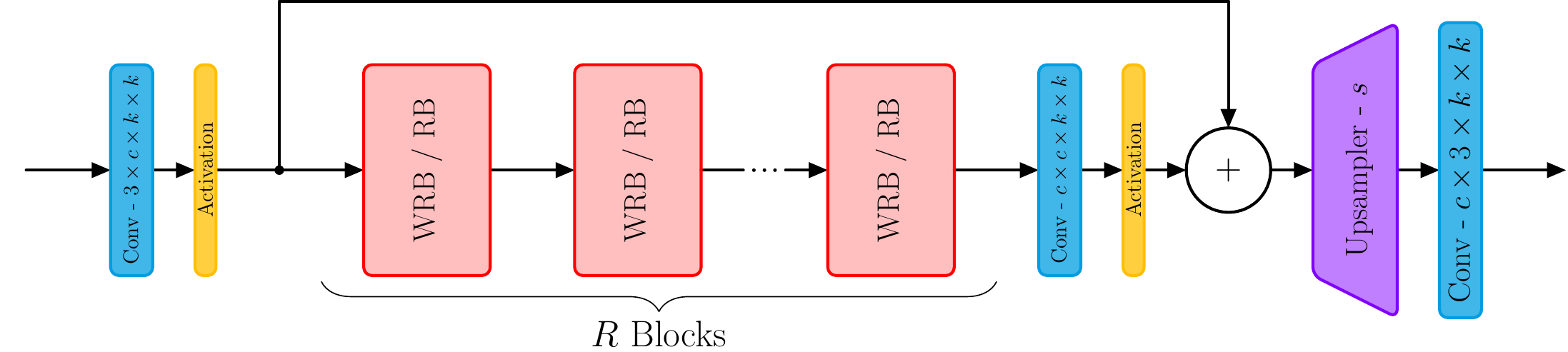} \vspace{-4pt}
	\caption{The model architecture for the super-resolution experiments. A shallow feature extractor layer is followed by a series of \layer[] blocks. The refined features are added to the initial extracted features to form an image in the desired resolution. \vspace{-6pt}}
	\label{fig:sr_arch}
\end{figure*}

\subsection{Single Image Super-Resolution (SISR)}
\label{ssec:sisr}

\subsubsection{Architecture}
\label{sssec:sisr_arch}

\begin{figure}
	\centering
	\includegraphics[width=0.42\textwidth]{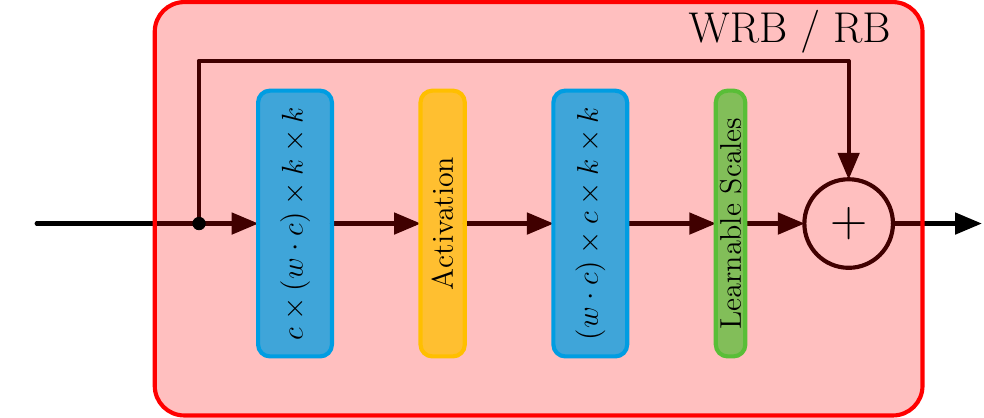}
	\caption{The structure of residual blocks (RB) and wide residual blocks (WRB). For WRB, $w>1$, while for RB, $w=1$.}
	\label{fig:resblock}
\end{figure}

The basic architecture chosen for the SISR task is the ResNet, which is widely used since the seminal paper \cite{ledig2017photo}. In this architecture, a single feature extraction layer is followed by a series of blocks for residual feature refinement. For simplicity, we selected residual blocks~\cite{he2016deep} for high-ordered neurons and wide residual blocks~\cite{zagoruyko2016wide} for first-order convolutions with scaled residuals \cite{szegedy2017inception} as our feature refinement blocks. The refined residual features are added back to the initial features, and the sum is processed by a feature upsampler module, which includes a convolutional layer, a non-linear activation function, and a PixelShuffler layer \cite{shi2016real}. This architecture is depicted in Fig. \ref{fig:sr_arch}, and the structure of residual and wide residual blocks is shown in Fig. \ref{fig:resblock}. Each learnable scalar layer for the channels is initialized as $0.1$.

\subsubsection{Training Details}
\label{sssec:sisr_train_Detail}

For the training models, we use DF2K dataset~\cite{lim2017enhanced}, which has more variety of images compared to DIV2K \cite{Agustsson_2017_CVPR_Workshops, Timofte_2017_CVPR_Workshops}. The models are trained on $64\times64$ patches scaled to the range $[-1, 1]$ with  batch size $25$ for $5\times10^{5}$ iterations to perform both $\times2$ and $\times4$ super-resolution. To enhance the training data, we apply random rotation, horizontal and vertical flip, and color channel shuffling as data augmentation. Additionally, we observed that adding a small amount of Gaussian noise during training improves the validation score of the network. Therefore, we add Gaussian noise with $40\text{ dB}$ SNR to the cropped patches. The model minimizes the loss function, proposed in \cite{barron2019general}, with parameters $\alpha=1.5$ and $c=2$. We employ the Adan optimizer \cite{xie2022adan} with an initial learning rate of $10^{-3}$ and utilize a cosine annealing scheduler \cite{loshchilov2016sgdr} to gradually decrease the learning rate until it reaches $10^{-6}$. The best model based on its validation PSNR on the DIV2K validation set is saved.

\subsubsection{\texorpdfstring{\neuron[]\ Configuration}{Paon Configuration}}
\label{sssec:neuron_settings}

We provide an analysis of some of the design choices, which were not studied in our earlier work~\cite{keles2024paon}. One of the key considerations is determining the~degree $[K/L]$ considering performance vs. model complexity. To this effect, we provide a comparison of degrees $[1/1]$, $[2/0]$ and $[2/1]$ using the same \shifter setting as in~\cite{keles2024paon}. The results presented in Table~\ref{tbl:mn_effect} show that although the best model is with degrees $[2/1]$, the second best model with $[1/1]$ demonstrates very similar performance while having more than $\%25$ fewer parameters. The table also indicates that, despite having the same number of parameters, the degree $[1/1]$ outperforms the degree $[2/0]$. We hypothesize that this difference in performance is due to the effective degree of approximation achieved by Eq.~\eqref{eq:pade_neuron_smooth}. For the $[1/1]$ configuration, the effective degree is $4$, whereas for $[2/0]$, it remains $2$. Consequently, we choose to proceed with the order $[1/1]$ for further experiments and evaluations.

\begin{table}[ht]
	\centering
	\caption{SISR $\times4$ experiments on the degree of \neuron[]. The~top row shows RGB-PSNR and Y-SSIM on DIV2K validation dataset. The bottom row is the number of parameters.}
	\resizebox{0.49\textwidth}{!}{
		\begin{tabular}{|c||c|c|c|}
			\hline
			$[K/L]$ & $[1/1]$ & $[2/0]$ & $[2/1]$ \\\hhline{|=||=|=|=|}
			PSNR / SSIM & $28.82/0.8179$ & $28.79/0.8170$ & $28.85/0.8187$ \\ 
			\# of parameters & $\approx 469$K & $\approx 469$K & $\approx 593$K \\\hline 
		\end{tabular}
	}
	\label{tbl:mn_effect}
\end{table}

Another key consideration is the \shifter configuration. Our previous work~\cite{keles2024paon} used only kernel-wise shifts. In this paper, we introduce element-wise shifts via offsets applied to input features, where offsets are determined through convolutional layers. Here, we evaluate the performance of the~\shifter with $1\times1$ and $3\times3$ offset calculation kernels with the goal of keeping the parameter count as low as possible. The~results are presented in Table~\ref{tbl:fullDeform_offsetKernel} show that applying element-wise shifts provides better performance compared to kernel-wise shifts. Moreover, using  deformable convolution with a $1\times1$ offset calculation kernel decreases the number of parameters compared to $3\times3$ offset calculation kernel without adversely affecting the PSNR performance. 

\begin{table}
	\centering
	\caption{SISR $\times4$ experiments on \shifter configuration. The~top line shows RGB-PSNR and Y-SSIM on DIV2K validation dataset. The bottom line is the number of parameters.} \vspace{-3pt}
	\resizebox{0.43\textwidth}{!}{
		\begin{tabular}{|c||c|c|}
			\hline
			Offset Kernel & $1\times1$ & $3\times3$ \\\hhline{|=||=|=|}
			\multirow{2}{*}{Kernel-wise shift} 
			& $28.81/0.8177$ & $28.80/0.8173$ \\ 
			& $\approx 441$K & $\approx 445$K \\\hline 
			\multirow{2}{*}{Element-wise shift} 
			& $28.84/0.8182$ & $28.83/0.8179$ \\ 
			& $\approx 445$K & $\approx 487$K \\\hline 
		\end{tabular}
	}
	\label{tbl:fullDeform_offsetKernel}
\end{table}

As a result, we have chosen to proceed with the degrees~$[1/1]$ and element-wise shifting using a $1\times1$ offset kernel, which provides a good  performance vs. parameter efficiency trade-off for the remainder of the experiments.

\subsubsection{Necessity of Singularity Prevention}
\label{ssec:need_for_sing_prevnt}

\begin{figure}
	\centering
	\subfloat{\label{subfig:11} \includegraphics[trim={0 0 0 0.85cm},clip,width=0.475\linewidth]{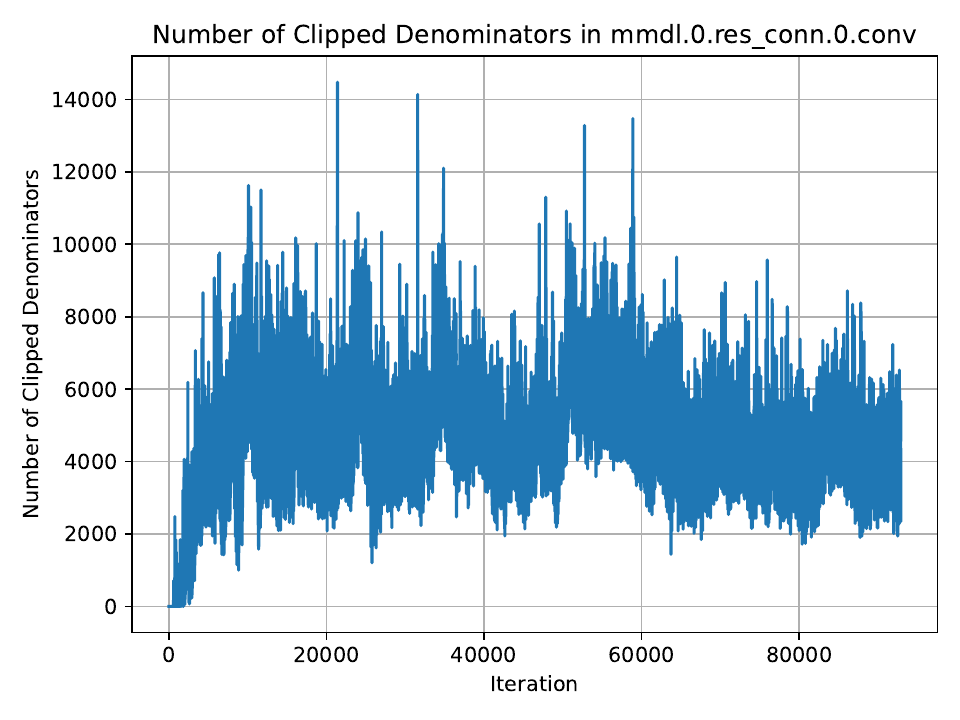}}
	\subfloat{\label{subfig:12} \includegraphics[trim={0 0 0 0.85cm},clip,width=0.475\linewidth]{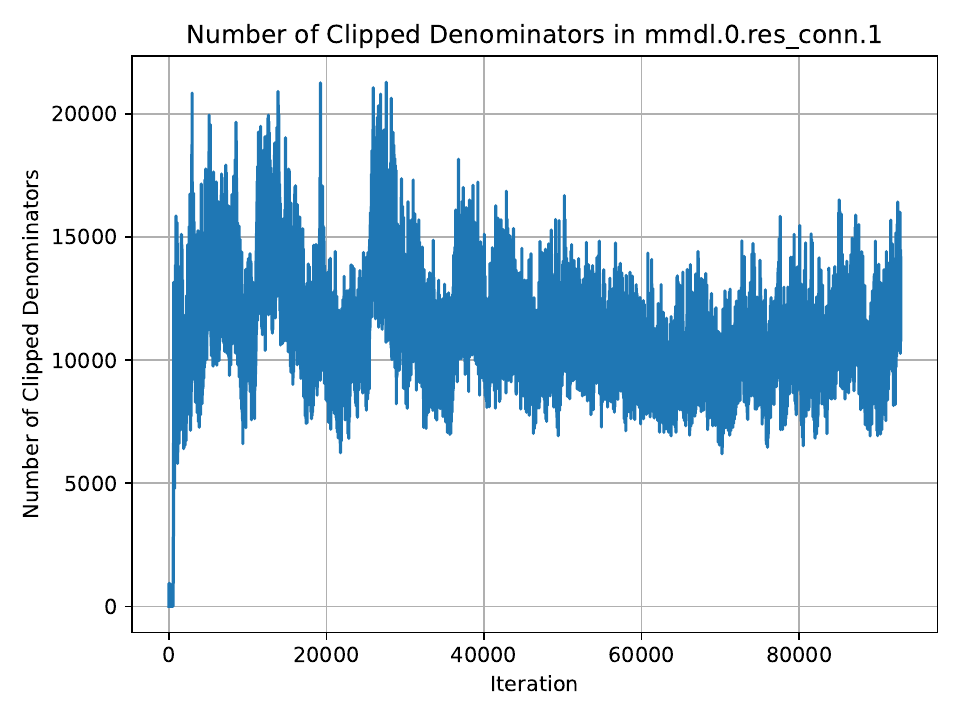}}\\
    \subfloat{\label{subfig:21} \includegraphics[trim={0 0 0 0.85cm},clip,width=0.475\linewidth]{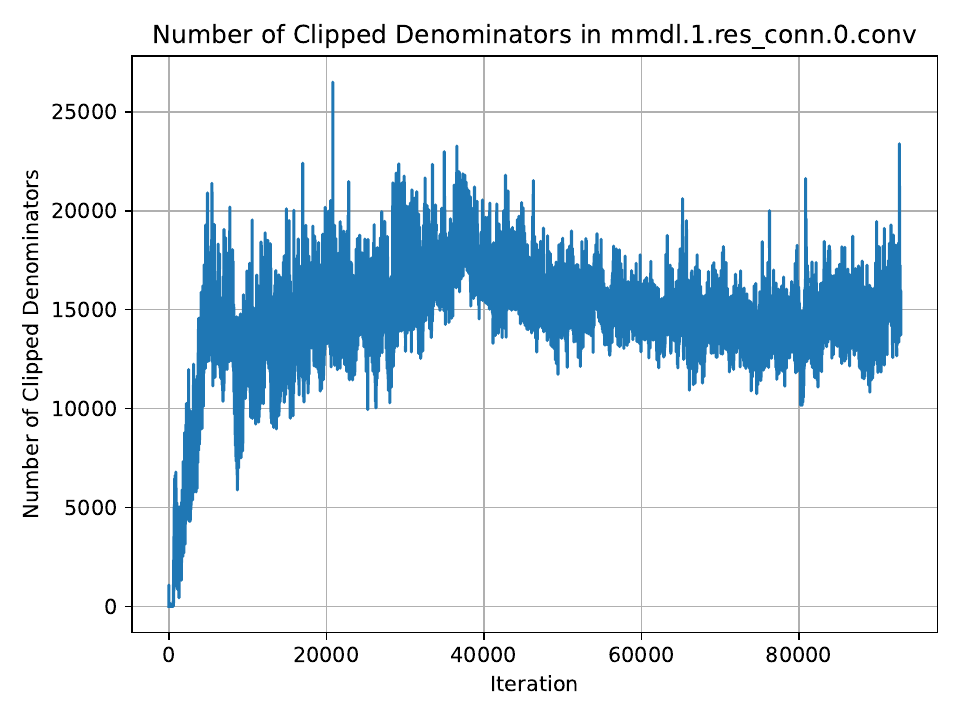}}
	\subfloat{\label{subfig:22} \includegraphics[trim={0 0 0 0.85cm},clip,width=0.475\linewidth]{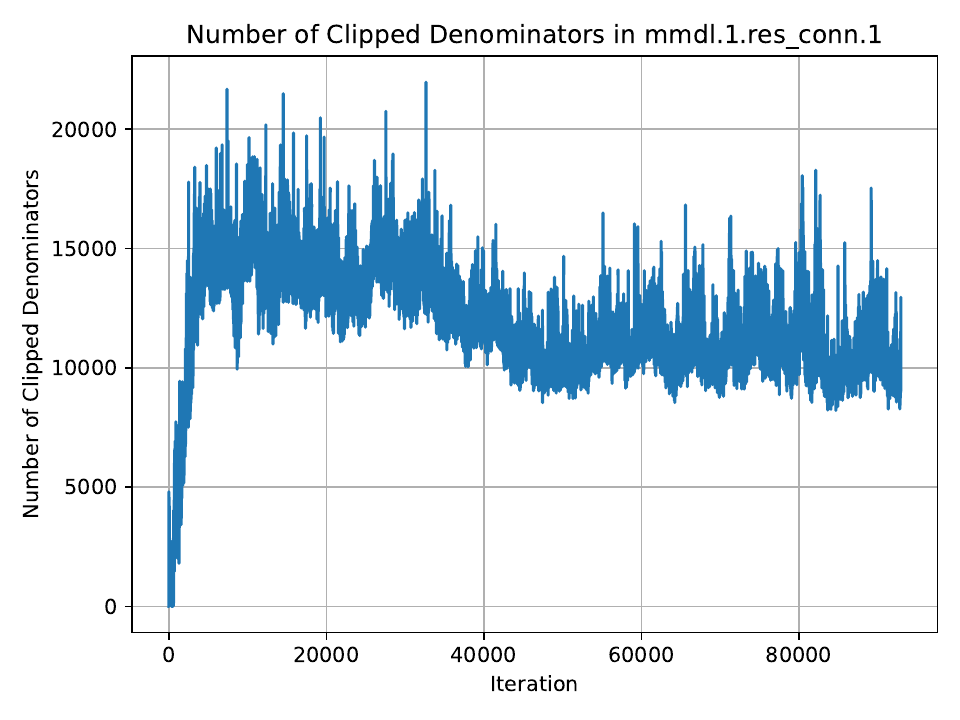}}\\
    \subfloat{\label{subfig:31} \includegraphics[trim={0 0 0 0.85cm},clip,width=0.475\linewidth]{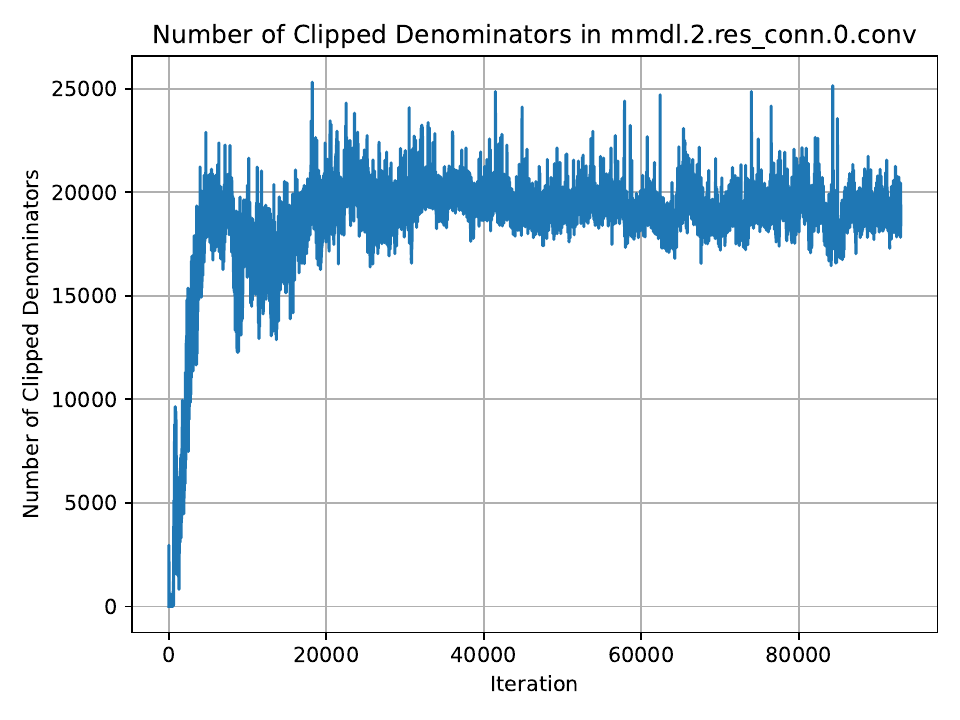}}
	\subfloat{\label{subfig:32} \includegraphics[trim={0 0 0 0.85cm},clip,width=0.475\linewidth]{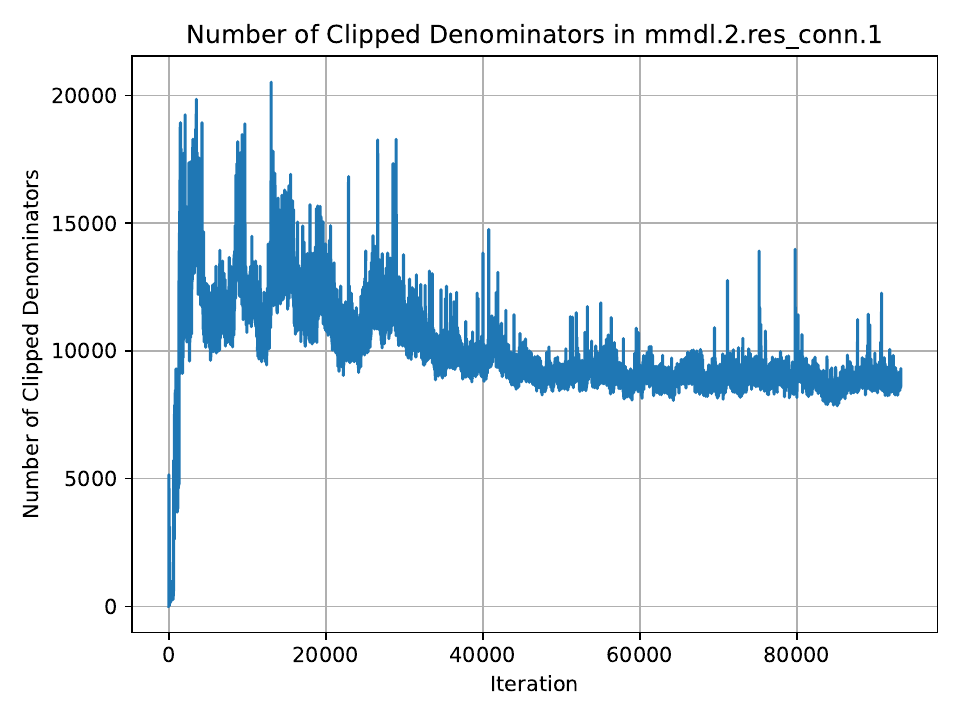}}
	\caption{Number of times the denominator $Q_{L}(n_1,n_2)$ is close to $0$ vs.  number of training iterations. The first, second, and third rows show the plots for the first, second and third residual blocks, respectively. The first and second columns correspond to the first and second layers in each residual block.}
	\label{fig:close_to_0}
\end{figure}

This subsection illustrates the need to employ the smoothed Pade approximation given by (\ref{eq:pade_neuron_smooth}) instead of (\ref{eq:vanilla_pade_neuron}). To demonstrate how often the~denominator $Q_{L}(n_1,n_2)$ in Eq. \eqref{eq:vanilla_pade_neuron} approaches zero during training when using the vanilla Padé approximation without stabilization, we provide plots of the denominator, $Q_{L}(n_1,n_2)$, and counts of the number of times $\lvert Q_{L}(n_1,n_2)\rvert$ falls below a threshold of $0.01$ for each layer of a model with $3$ residual blocks (two layers in each residual block) during each iteration of training. The horizontal axis shows the~number of iterations during training. We note that if we let the denominator to approach zero, we encounter instabilities that frequently lead to early termination of training. The plots given in Fig. \ref{fig:close_to_0} illustrate that the denominator of the Padé approximant without stabilization approaches zero between $5,000$ and $20,000$ times for each layer in each residual block, which clearly demonstrates the need for the proposed smoothed Padé approximation (\neuron[]$^S$) given by Eq. \eqref{eq:pade_neuron_smooth} to guarantee stability and performance of the Padé neurons.

\subsubsection{Comparison of Model Performance}
\label{sssec:sisr_results}

We ran tests on standard datasets used in the SISR literature, which include BSD100~\cite{martin2001database}, Manga109~\cite{matsui2017sketch}, Set5 \cite{bevilacqua2012low}, Set14 \cite{zeyde2012single}, Urban100~\cite{huang2015single}, and DIV2K validation dataset. The performance of all models is evaluated using the peak signal-to-noise ratio (PSNR), structural similarity index (SSIM), and learned perceptual image patch similarity (LPIPS) metrics. The PSNR~\cite{keles2021computation} is calculated on RGB images, while SSIM~\cite{wang2004image} is computed on the Y channel of images. LPIPS \cite{zhang2018unreasonable} results are reported from both AlexNet~\cite{krizhevsky2012imagenet} and VGG \cite{simonyan2014very} networks.

We evaluate the performance of \network[] with and without an external activation vs. different neuron models using the same network architecture, including wide residual network (ResNet), which is composed of classic convolutional layers with \mainact\ activation, \pau-Net with classic convolutional layers and \pau\ activation, deformable convolution networks (DCN $k\times k$) using various offset convolution kernel sizes, a SelfONN, a SuperONN. In all models, we maintain the same number of $[1/0]$ convolution layers in the initial feature extractor, the feature refinement endpoint, and the final image constructor (including the upsampler and final layer). Different neuron models are used only in the residual feature refinement layers to ensure fair comparison with similar number of parameters across different architectures. The architecture details for all of the compared models are presented in Table~\ref{tbl:archs_overview}.

\begin{table}[th!]
	\centering
	\caption{Configuration of compared models. For \pau-Net,  the degree of \pau\ activation is shown. \textquotedblleft RB\textquotedblright\ and \textquotedblleft WRB\textquotedblright\ denote residual block (RB) and wide RB, respectively. \textquotedblleft Deform.\textquotedblright\ means deformable, and \textquotedblleft KW\textquotedblright\ is kernel-wise shift. \network[] has \mainact\ activation, \network[-ID] has no activation.}
	\resizebox{0.49\textwidth}{!}{
		\begin{tabular}{|c||c|c|c|c|c|}
			\hline
			\multicolumn{1}{|c||}{} & ResNet & DCN $k\times k$ & \pau-Net & Self/SuperONN & \network[](-ID) \\\hhline{|=||=|=|=|=|=|}
			$[K/L]$ & $[1/0]$ & $[1/0]$ & $[7/6]$ & $[2/0]$ & $[1/1]$ \\\hline
			Activation & \mainact & \mainact & \pau & \mainact & \mainact/-- \\\hline
			Blocks, $R$ & $3$ & $3$ & $3$ & $3$ & $3$ \\\hline
			Type ($w$) & WRB ($2$) & WRB ($2$) & WRB ($2$) & RB ($1$) & RB ($1$) \\\hline
			Channels & $48$ & $48$ & $48$ & $48$ & $48$ \\\hline
			Strategy & -- & Deform. & -- & --/KW & Deform. \\\hline
			Shift Kernel & -- & $k\times k$ & -- & -- & $1\times1$ \\\hline
		\end{tabular}
	}
	\label{tbl:archs_overview}
\end{table}

Quantitative comparison of \network[] and \network[-ID] (no activation) vs. competing models in Table~\ref{tbl:archs_overview} in terms of fidelity metrics (PSNR and SSIM) and LPIPS are shown in Table \ref{tbl:quant_results}. Comparison of the last two rows show that \network[-ID], which uses \neuron[s] without any fixed activation performs better than \neuron[s] using GeLU activation. Furthermore, \network[-ID] consistently achieves the best fidelity performance across all datasets. Comparison of \network[-ID] with SelfONN and SuperONN clearly demonstrates the superior function approximation capability of \neuron[s] compared to generative neurons \emph{without  the need for additional non-linear activation}. Table \ref{tbl:quant_results} also validates that $1\times1$ offset kernel performs better than $3\times3$ kernel even within deformable convolution networks.

In order to explore whether we can increase the performance of \network[] within the same parameter budget, we use two $\times2$ PixelShuffler layers with shared weights instead of two independent $\times2$ PixelShuffler layers. Comparison of results presented in   Table~\ref{tbl:quant_results} vs. Table~\ref{tbl:results360kshared} shows that there is a small performance loss due to using shared $\times2$ PixelShuffler layers. However, if we add one more residual block to each model using the parameters saved by using shared weights in the two PixelShuffler layers, the results in Table~\ref{tbl:results450kshared} indicate that we gain more than what we lose in all models within the same approximately $450$K parameter budget.

\begin{table*}[h!]
	\centering
	\caption{Quantitative comparison for $\times 4$ SISR task. The number below the model shows the number of parameters. The top row in each cell shows PSNR($\uparrow$) and SSIM($\uparrow$), and the bottom row shows LPIPS($\downarrow$) based on AlexNet / VGGNet, respectively. \network[-ID] does not use any fixed activation. The best and second best scores are shown in \f{red} and \s{blue}, respectively.}
		\begin{tabular}{|c||c|c|c|c|c|} 
			\cline{2-6}
        \hline
			Test Set & BSD100 & Manga109 & Set5 & Set14 & Urban100  \\\hhline{|=||=|=|=|=|=|}
			ResNet
			& $26.10/0.7123$ & $28.02/0.8904$ & $29.76/0.8768$ & $26.21/0.7585$ & $24.19/0.7567$ \\
			
			$\approx 440$K & $0.3892/0.3447$ & $0.1186/0.1739$ & $0.1815/0.2190$ & $0.2956/0.3105$ & $0.2572/0.2978$ \\\hline
			DCN $1\times1$
			& $26.14/0.7133$ & $28.16/0.8923$ & $\s{29.90}/0.8781$ & $26.34/0.7595$ & $24.30/0.7604$ \\
			
			$\approx 447$K & $0.3870/\s{0.3434}$ & $0.1160/0.1719$ & $0.1803/0.2171$ & $0.2950/\s{0.3072}$ & $0.2522/0.2942$ \\\hline
			DCN $3\times3$
			& $26.14/0.7132$ & $28.13/0.8917$ & $29.87/0.8779$ & $26.33/0.7593$ & $24.27/0.7594$ \\
			
			$\approx 509$K & $\f{0.3864}/0.3435$ & $0.1162/0.1724$ & $0.1802/0.2174$ & $0.2950/0.3083$ & $0.2535/0.2952$ \\\hline
			\pau-Net
			& $26.08/0.7116$ & $27.98/0.8894$ & $29.73/0.8758$ & $26.20/0.7576$ & $24.15/0.7551$ \\
			
			$\approx 440$K & $0.3908/0.3435$ & $0.1188/\f{0.1711}$ & $0.1805/\s{0.2167}$ & $0.2971/0.3091$ & $0.2600/0.2977$ \\\hline
			SelfONN
			& $26.11/0.7123$ & $28.08/0.8908$ & $29.80/0.8770$ & $26.33/0.7586$ & $24.22/0.7573$ \\
			
			$\approx 440$K & $0.3896/0.3452$ & $0.1178/\s{0.1717}$ & $0.1804/0.2181$ & $0.2964/0.3090$ & $0.2567/0.2956$ \\\hline
			SuperONN
			& $26.11/0.7122$ & $28.05/0.8902$ & $29.81/0.8768$ & $26.30/0.7583$ & $24.22/0.7571$ \\
			
			$\approx 440$K & $0.3892/0.3449$ & $0.1174/0.1720$ & $0.1808/0.2190$ & $0.2962/0.3096$ & $0.2562/0.2956$ \\\hhline{|=||=|=|=|=|=|} 
			\network[]
			& $\s{26.15}/\s{0.7139}$ & $\s{28.25}/\s{0.8935}$ & $\f{29.96}/\s{0.8792}$ & $\s{26.37}/\s{0.7604}$ & $\s{24.35}/\s{0.7622}$ \\
			
			$\approx 445$K & $\s{0.3868}/\s{0.3434}$ & $\s{0.1146}/\s{0.1717}$ & $\f{0.1797}/\f{0.2163}$ & $\f{0.2943}/0.3073$ & $\s{0.2503}/\s{0.2933}$ \\\hline
			\network[-ID]
			& $\f{26.17}/\f{0.7144}$ & $\f{28.28}/\f{0.8940}$ & $\f{29.96}/\f{0.8793}$ & $\f{26.39}/\f{0.7607}$ & $\f{24.38}/\f{0.7636}$ \\
			
			$\approx 445$K & $\f{0.3864}/\f{0.3428}$ & $\f{0.1136}/\s{0.1717}$ & $\s{0.1799}/0.2171$ & $\s{0.2946}/\f{0.3064}$ & $\f{0.2476}/\f{0.2924}$ \\\hline
		\end{tabular}
	\label{tbl:quant_results}
\end{table*}
\begin{table*}[h]
	\centering
	\caption{Quantitative comparison around $360$K parameters with shared pixel shuffler for $\times 4$ SISR. The top two scores in each cell are PSNR($\uparrow$) and SSIM($\uparrow$), and the bottom two are LPIPS($\downarrow$) based on AlexNet / VGGNet, respectively. }
		\begin{tabular}{|c||c|c|c|c|c|} 
			\hline
			Test Set & BSD100 & Manga109 & Set5 & Set14 & Urban100 \\\hhline{|=||=|=|=|=|=|}
			ResNet
			& $26.08/0.7114$ & $27.93/0.8887$ & $29.70/0.8757$ & $26.17/0.7579$ & $24.13/0.7546$ \\ 
			
			$\approx 357$K & $0.3914/0.3455$ & $0.1202/0.1737$ & $0.1827/0.2189$ & $0.2977/0.3103$ & $0.2618/0.2996$ \\\hline
			DCN $1\times1$
			& $\s{26.13}/\s{0.7127}$ & $28.08/\s{0.8910}$ & $29.81/0.8769$ & $\s{26.33}/\s{0.7592}$ & $24.25/0.7584$ \\ 
			
			$\approx 363$K & $0.3890/0.3444$ & $0.1177/0.1727$ & $\s{0.1807}/0.2189$ & $0.2963/\s{0.3081}$ & $0.2563/0.2968$ \\\hline
			DCN $3\times3$
			& $\s{26.13}/0.7126$ & $\s{28.09}/\s{0.8910}$ & $\f{29.89}/\s{0.8778}$ & $\f{26.34}/\s{0.7592}$ & $\s{24.26}/\s{0.7585}$ \\ 
			
			$\approx 426$K & $\s{0.3889}/\s{0.3443}$ & $\s{0.1175}/0.1723$ & $\f{0.1805}/\s{0.2178}$ & $\s{0.2954}/0.3083$ & $\s{0.2559}/\s{0.2963}$ \\\hline
			SelfONN
			& $26.10/0.7118$ & $28.03/0.8898$ & $29.84/0.8772$ & $26.29/0.7585$ & $24.18/0.7557$ \\ 
			
			$\approx 357$K & $0.3916/0.3457$ & $0.1193/\f{0.1714}$ & $\f{0.1805}/\f{0.2171}$ & $0.2983/0.3103$ & $0.2608/0.2977$ \\\hline
			SuperONN
			& $26.10/0.7117$ & $27.96/0.8887$ & $29.78/0.8760$ & $26.27/0.7577$ & $24.17/0.7555$ \\ 
			
			$\approx 357$K & $0.3918/0.3451$ & $0.1192/0.1721$ & $0.1819/0.2180$ & $0.2978/0.3103$ & $0.2598/0.2972$ \\\hhline{|=||=|=|=|=|=|} 
			\network[-ID]
			& $\f{26.15}/\f{0.7131}$ & $\f{28.21}/\f{0.8930}$ & $\s{29.86}/\f{0.8779}$ & $\f{26.34}/\f{0.7599}$ & $\f{24.32}/\f{0.7615}$ \\ 
			
			$\approx 362$K & $\f{0.3868}/\f{0.3437}$ & $\f{0.1156}/\s{0.1716}$ & $0.1809/\s{0.2178}$ & $\f{0.2946}/\f{0.3078}$ & $\f{0.2524}/\f{0.2942}$ \\\hline
		\end{tabular}
	\label{tbl:results360kshared}
\end{table*}
\begin{table*}[h!]
	\centering
	\caption{Quantitative comparison around $450$K parameters with shared pixel shuffler and $4$ residual blocks for $\times4$ SISR. The top row in each cell show PSNR($\uparrow$) and SSIM($\uparrow$), and the bottom row  LPIPS($\downarrow$) based on AlexNet / VGGNet, respectively. }
		\begin{tabular}{|c||c|c|c|c|c|} 
			\hline
			Test Set & BSD100 & Manga109 & Set5 & Set14 & Urban100 \\\hhline{|=||=|=|=|=|=|}
			ResNet
			& $26.13/0.7132$ & $28.13/0.8919$ & $29.79/0.8777$ & $26.26/0.7595$ & $24.26/0.7594$ \\ 
			
			$\approx 440$K & $0.3894/0.3434$ & $0.1176/0.1722$ & $0.1809/0.2181$ & $0.2957/0.3091$ & $0.2559/0.2958$ \\\hline
			DCN $1\times1$
			& $\s{26.16}/\s{0.7143}$ & $\s{28.26}/\s{0.8934}$ & $\s{29.95}/\s{0.8789}$ & $\s{26.39}/\s{0.7609}$ & $\s{24.35}/\s{0.7622}$ \\ 
			
			$\approx 449$K & $\s{0.3873}/\f{0.3428}$ & $\s{0.1155}/\f{0.1698}$ & $\s{0.1793}/\f{0.2163}$ & $\s{0.2940}/\s{0.3070}$ & $\s{0.2522}/\f{0.2925}$ \\\hline
			DCN $3\times3$
			& $\s{26.16}/0.7140$ & $28.22/0.8930$ & $29.92/0.8788$ & $26.36/0.7605$ & $\s{24.35}/\s{0.7622}$ \\
			
			$\approx 532$K & $0.3875/0.3439$ & $0.1160/0.1718$ & $0.1799/\s{0.2172}$ & $0.2955/0.3078$ & $0.2524/0.2940$ \\\hline
			SelfONN
			& $26.14/0.7132$ & $28.15/0.8918$ & $29.83/0.8776$ & $26.35/0.7596$ & $24.29/0.7600$ \\ 
			
			$\approx 440$K & $0.3902/0.3445$ & $0.1174/\s{0.1704}$ & $0.1816/0.2176$ & $0.2966/0.3087$ & $0.2553/0.2939$ \\\hline
			SuperONN
			& $26.13/0.7128$ & $28.09/0.8910$ & $29.84/0.8775$ & $26.32/0.7591$ & $24.27/0.7595$ \\ 
			
			$\approx 440$K & $0.3902/0.3451$ & $0.1172/0.1719$ & $0.1823/0.2180$ & $0.2968/0.3094$ & $0.2552/0.2945$ \\\hhline{|=||=|=|=|=|=|} 
			\network[-ID]
			& $\f{26.18}/\f{0.7149}$ & $\f{28.32}/\f{0.8944}$ & $\f{29.99}/\f{0.8799}$ & $\f{26.41}/\f{0.7611}$ & $\f{24.41}/\f{0.7645}$ \\ 
			
			$\approx 447$K & $\f{0.3857}/\s{0.3430}$ & $\f{0.1145}/0.1708$ & $\f{0.1791}/0.2176$ & $\f{0.2933}/\f{0.3064}$ & $\f{0.2488}/\s{0.2927}$ \\\hline
		\end{tabular}
	\label{tbl:results450kshared}
\end{table*}

\begin{figure*}
	\centering
	\begin{tabular}{c}
		\begin{tikzpicture}
			\node[anchor=south west,inner sep=1] (img) at (0,0) {\includegraphics[width=0.38\textwidth,trim={0 2 0 2},clip]{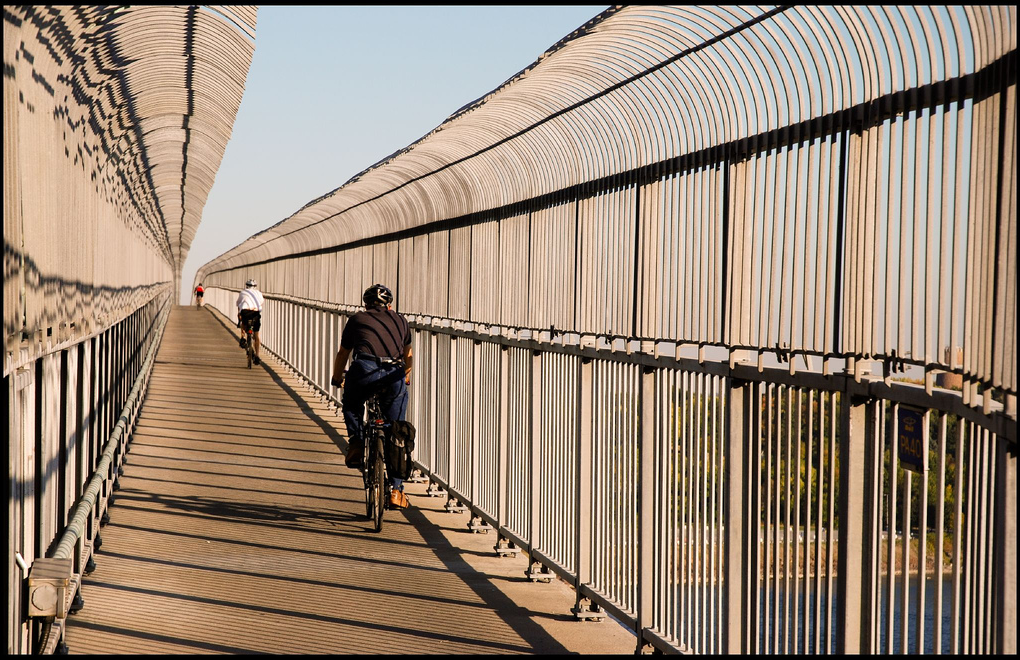}};
			\begin{scope}[x={(img.south east)},y={(img.north west)}]
				\draw[green,thick] (0.6598,0.8963) rectangle (0.7382,0.7744); 
			\end{scope}
		\end{tikzpicture}
	\end{tabular}
	\begin{tabular}{cccc}
		\includegraphics[width=0.119\textwidth]{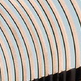} &
		\includegraphics[width=0.119\textwidth]{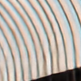} & 
		\includegraphics[width=0.119\textwidth]{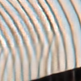} & 
		\includegraphics[width=0.119\textwidth]{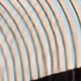} \\
		\includegraphics[width=0.119\textwidth]{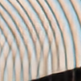} &
		\includegraphics[width=0.119\textwidth]{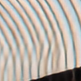} & 
		\includegraphics[width=0.119\textwidth]{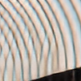} & 
		\includegraphics[width=0.119\textwidth]{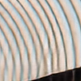}
	\end{tabular}
	\begin{tabular}{c}
		\begin{tikzpicture}
			\node[anchor=south west,inner sep=1] (img) at (0,0) {\includegraphics[width=0.38\textwidth,trim={0 50 0 50},clip]{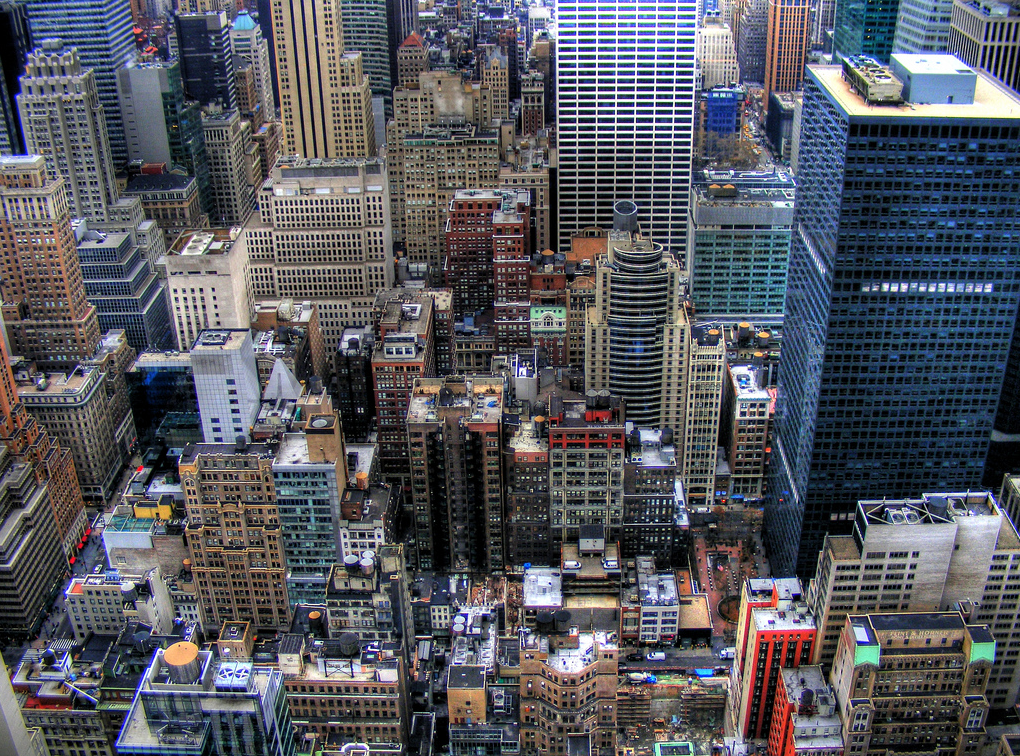}};
			\begin{scope}[x={(img.south east)},y={(img.north west)}]
				\draw[green,thick] (0.007,0.2607) rectangle (0.0794,0.1387); 
			\end{scope}
		\end{tikzpicture}
	\end{tabular}
	\begin{tabular}{cccc}
		\includegraphics[width=0.119\textwidth]{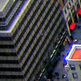} &
		\includegraphics[width=0.119\textwidth]{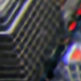} & 
		\includegraphics[width=0.119\textwidth]{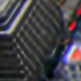} & 
		\includegraphics[width=0.119\textwidth]{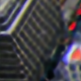} \\
		\includegraphics[width=0.119\textwidth]{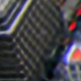} &
		\includegraphics[width=0.119\textwidth]{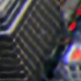} & 
		\includegraphics[width=0.119\textwidth]{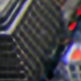} & 
		\includegraphics[width=0.119\textwidth]{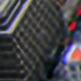}
	\end{tabular}
	\begin{tabular}{c}
		\begin{tikzpicture}
			\node[anchor=south west,inner sep=1] (img) at (0,0) {\includegraphics[width=0.38\textwidth,trim={0 8 0 8},clip]{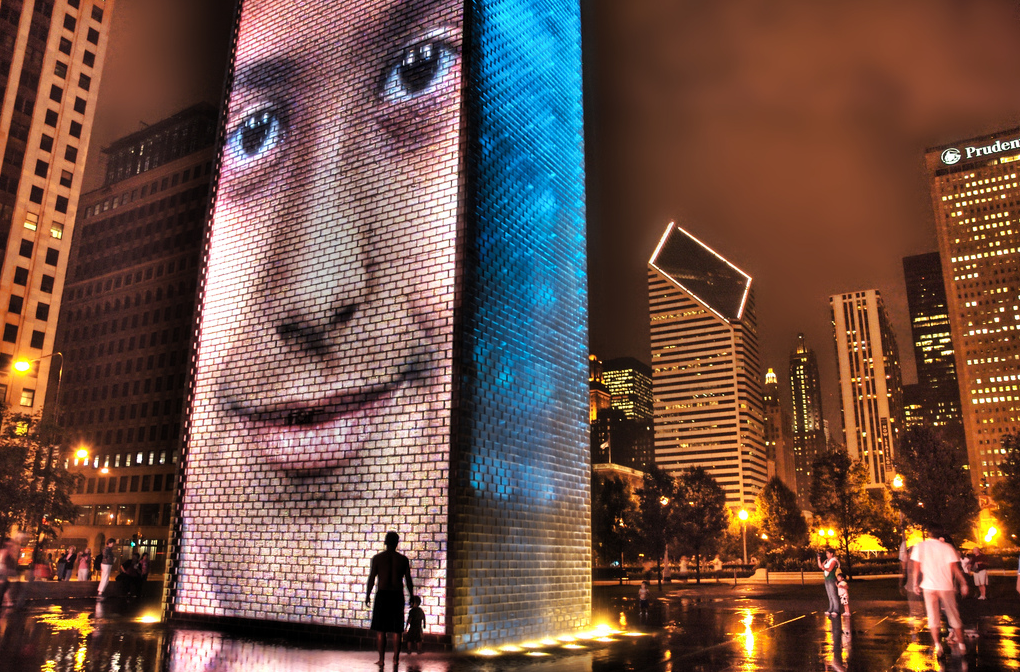}};
			\begin{scope}[x={(img.south east)},y={(img.north west)}]
				\draw[green,thick] (0.6127,0.5472) rectangle (0.7598,0.3229); 
			\end{scope}
		\end{tikzpicture}
	\end{tabular}
	\begin{tabular}{cccc}
		\includegraphics[width=0.119\textwidth]{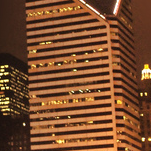} &
		\includegraphics[width=0.119\textwidth]{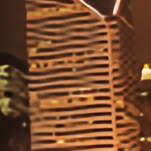} & 
		\includegraphics[width=0.119\textwidth]{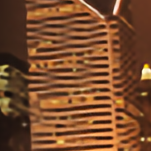} & 
		\includegraphics[width=0.119\textwidth]{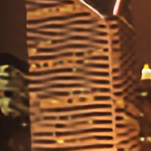} \\
		\includegraphics[width=0.119\textwidth]{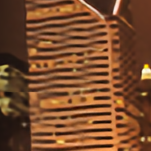} &
		\includegraphics[width=0.119\textwidth]{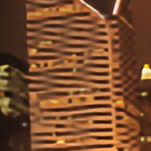} & 
		\includegraphics[width=0.119\textwidth]{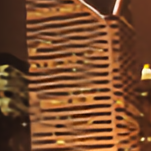} & 
		\includegraphics[width=0.119\textwidth]{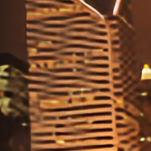}
	\end{tabular}
	\caption{Visual comparisons on \texttt{img\_024}, \texttt{img\_073} and \texttt{img\_076} from Urban100 dataset for $\times4$ SR. Crop-outs from left to right, top row: ground truth, \network[], SelfONN, SuperONN, bottom row: ResNet, \pau-Net, DCN $1\times1$ DCN, $3\times3$.}
	\label{fig:visual_comp}
\end{figure*}

Qualitative (visual) comparisons are presented in Fig. \ref{fig:visual_comp}. These results clearly indicate that \neuron[] exhibits superior performance compared to its competitors in terms of fidelity. For instance, the high-frequency patch on the top image (\texttt{img\_024.png}) is reconstructed best by the \network[], nearly without aliasing, whereas other methods introduce aliasing artifacts. This superior performance is also observed in other images: In the middle image (\texttt{img\_073.png}), the shown crop has the least amount of aliasing artifacts in the output of \network[]. For the bottom image (\texttt{img\_076.png}), the building stripes are mostly correctly oriented in the output of \network[]. These qualitative results confirm the quantitative findings, demonstrating that \network[] offers superior performance in preserving high-frequency details and structures in the image. This effectiveness is attributed to the superior representation capabilities of the proposed \neuron[s].
\vspace{-6pt}

\subsection{Image Compression}
\label{subsec:imgcomp}

\subsubsection{Architecture}
\label{sssec:imgcomp_arch}

We have chosen two popular image compression architectures, the joint autoregressive and hierarchical priors \cite{minnen2018joint} and ELIC \cite{he2022elic}, to show that replacing convolutional layers with \layer[s] improve performance. In our first model, called MBT-\neuron[], we replace all convolutional layers in MBT-2018~\cite{minnen2018joint} with  \layer[s] degree $[1/1]$ in the encoder and decoder. In our second model, ELIC-\neuron[], similar to the~approach outlined in \cite{luo2024rethinking}, we only replace convolutional layers in the~decoder, spatial context, and channel context model in ELIC \cite{he2022elic} with \layer[s] degree $[1/1]$. Moreover, in ELIC-\layer[], we reduce the number of residual bottleneck blocks from $3$ to $1$. In both architectures, the upsampling layers are implemented via transposed convolutions. Hence, we adopt kernel-wise shifting strategy instead of element-wise shifting.

\subsubsection{Training Details}
\label{sssec:imgcomp_train_Detail}

We combined selected images from ImageNet \cite{deng2009imagenet}, DF2K \cite{lim2017enhanced}, COCO 2017 \cite{lin2014microsoft}, and CLIC training dataset \cite{toderici2020workshop}, forming a dataset comprising over $100$K images. We conduct image compression experiments using the~codebase provided by \cite{jiang2023mlicpp}. In each experiment, $256\times256$ crops are taken at random, and a batch size of $24$ is used. Models are trained for $600$ epochs with an initial learning rate of $10^{-4}$ using six values of $\lambda = \{0.0018, 0.0035, 0.0067, 0.0130, 0.0250, 0.0483\}$ representing different rate-distortion (RD) trade-off points. The learning rate is reduced by a factor of $10$ at epochs $450$ and $550$ to fine-tune the model performance as training progresses. Additionally, we apply gradient clipping to stabilize the training process by limiting the maximum norm of the gradients to $1$. For the joint autoregressive model, the number of channels is set to $M=192$ and $N=192$ for the first four values of $\lambda$, and increased to $M=320$ and $N=192$ for the last two values. For ELIC, $M=320$ and $N=192$ for all $\lambda$ values.

\subsubsection{Comparison of Model Performance}
\label{sssec:imgcomp_results}

\begin{figure*}
	\centering
	\subfloat[]{\label{subfig:comp_mbt_rd}%
        \includegraphics[trim={0.55cm 0.3cm 1.55cm 0.8cm},clip,width=0.4\linewidth]{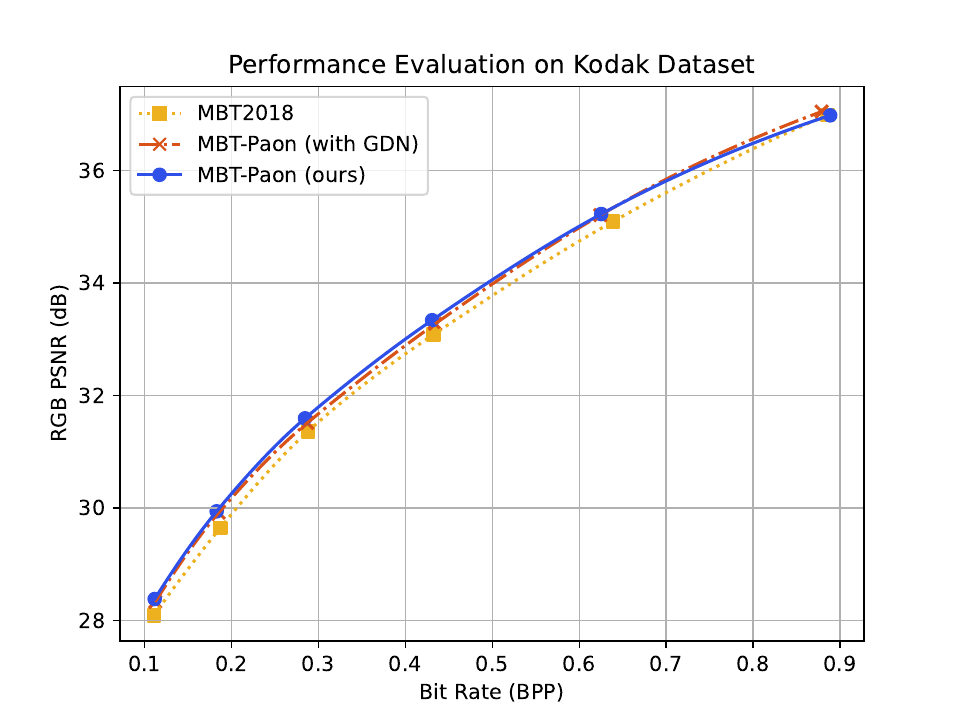}
    }
	\subfloat[]{\label{subfig:comp_mbt_saving}%
        \includegraphics[trim={1cm 4.2cm 1cm 4cm},clip,width=0.297\linewidth]{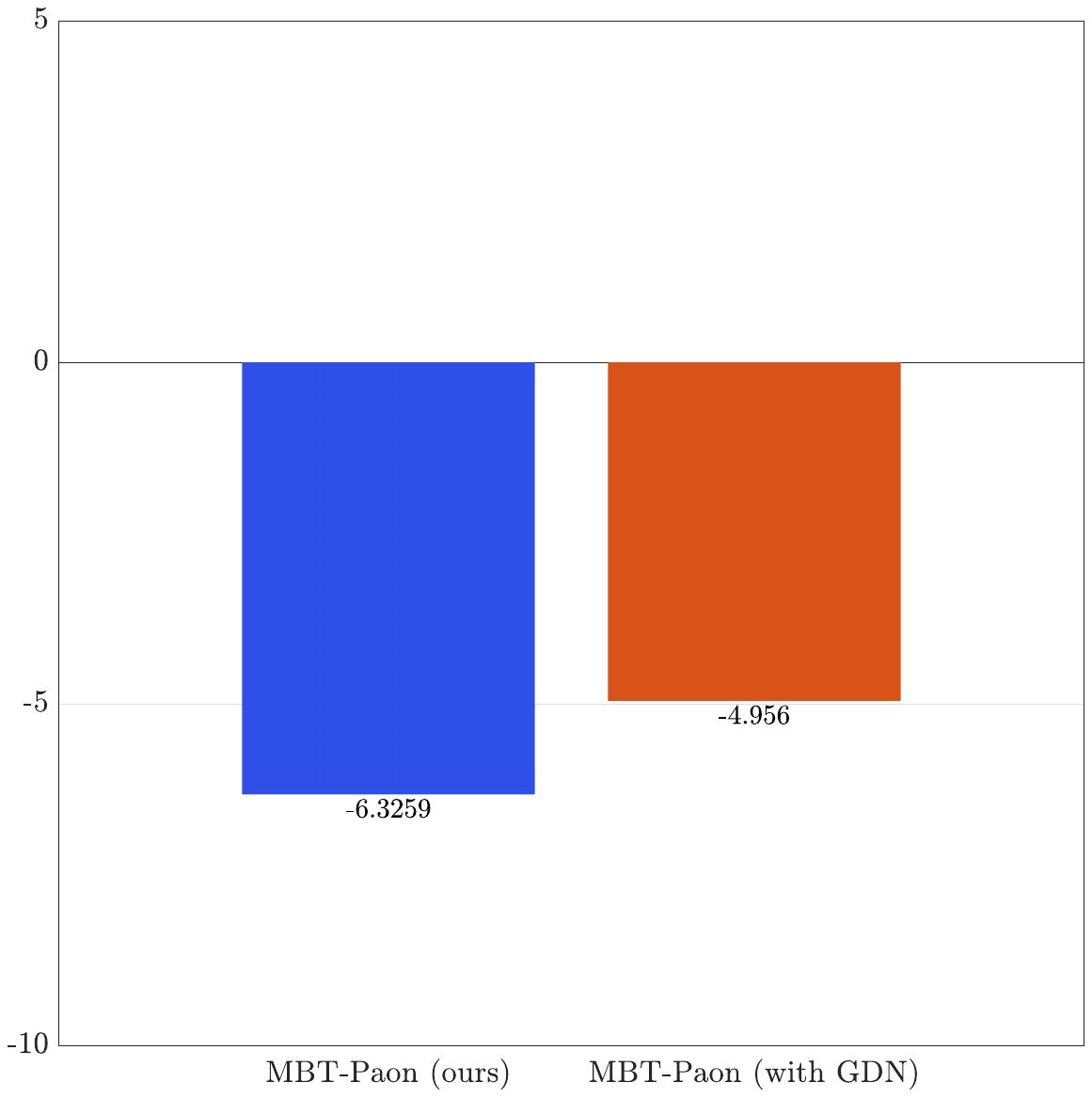}
    } 
	\caption{Comparison of the MBT-\neuron[] model vs. the benchmark MBT-2018. \protect\subref{subfig:comp_mbt_rd} RD curves of MBT-\neuron[] (with and without GDN) and the anchor Minnen et al. \cite{minnen2018joint}. \protect\subref{subfig:comp_mbt_saving} Average percent BD-rate savings for RGB PSNR with respect to the anchor model~\cite{minnen2018joint}.
    Observe that removal of GDN layers in MBT-\neuron[] not only results in compute savings but also in performance improvement.}
	\label{fig:comp_mbt_rd_and_saving}
\end{figure*}

Figure \ref{fig:comp_mbt_rd_and_saving} shows the~comparison of MBT-\neuron[] vs. MBT-2018~\cite{minnen2018joint}. The RD curve of MBT-2018 is taken from the CompressAI benchmark~\cite{begaint2020compressai}. In Figure \ref{subfig:comp_mbt_rd}, RD curves for MBT-\neuron[] with and without GDN layers both surpass that of the original MBT-2018 model. Notably, MBT-\neuron[] without GDN layers not only provides computational savings but also results in performance improvement over MBT-\neuron[] with GDN layers. The superiority of MBT-\neuron[] can also be seen from Figure~\ref{subfig:comp_mbt_saving}, showing the BD-rate \cite{bjontegaard2001calculation} improvements. Observe that MBT-\neuron[] without GDN layers relying only on the non-linear power of \layer[s] saves more than $6\%$ bit rate compared to the off-the-shelf benchmark model. These results indicate that simply substituting the common convolutional layers with \layer[s] brings a performance improvement without any bells and whistles. 

\begin{figure*}
	\centering
	\subfloat[]{\label{subfig:comp_elic_rd}%
        \includegraphics[trim={0.55cm 0.3cm 1.55cm 0.8cm},clip,width=0.4\linewidth]{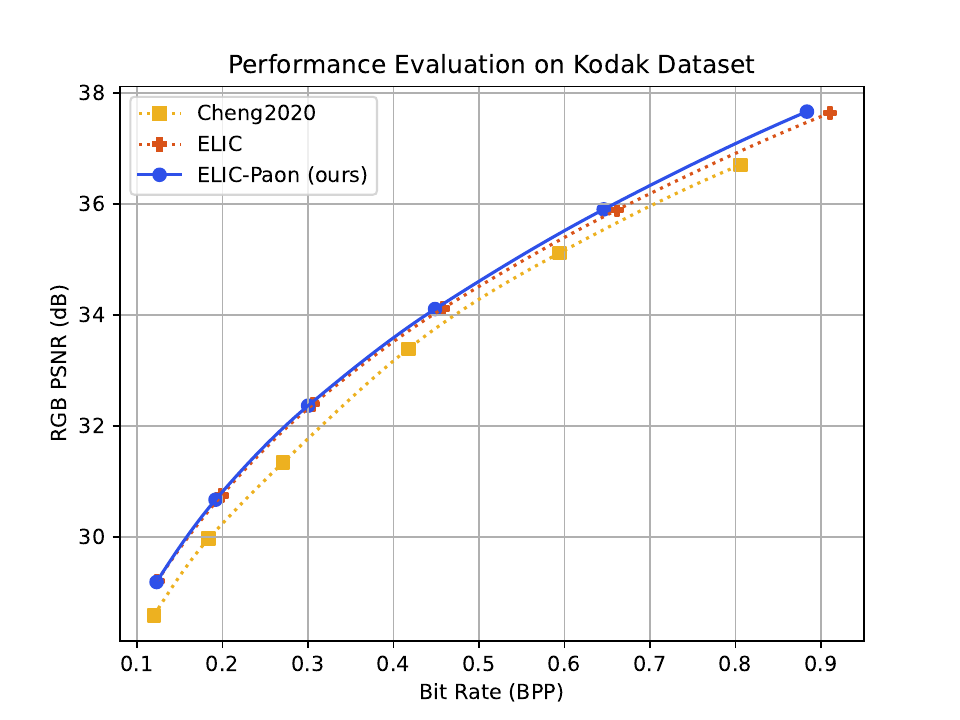}
    }
	\subfloat[]{\label{subfig:comp_elic_saving}%
        \includegraphics[trim={1cm 4.2cm 1cm 4cm},clip,width=0.297\linewidth]{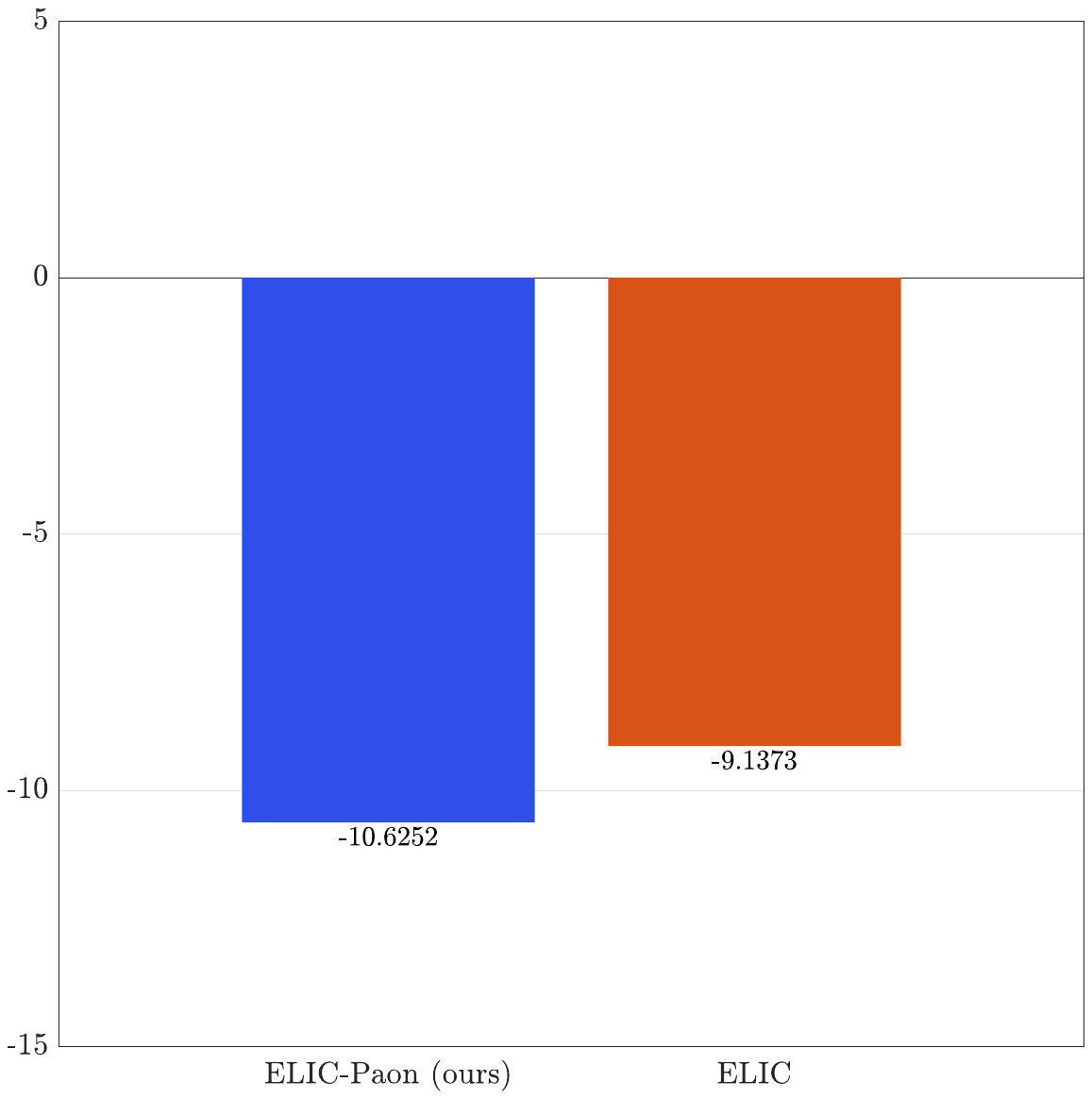}
    } 
	\caption{Comparison of the ELIC-Paon model with only one-third RBB vs. original ELIC. \protect\subref{subfig:comp_elic_rd} RD curves of ELIC-Paon, ELIC, and the anchor Cheng et al. \cite{cheng2020learned}. \protect\subref{subfig:comp_elic_saving} Average percent BD-rate savings for RGB PSNR with respect to the anchor model~\cite{cheng2020learned}.}
	\label{fig:comp_elic_rd_and_saving}
\end{figure*}

\begin{figure*}
	\centering
	\begin{tabular}{c}
		\includegraphics[trim={0 1.25cm 0 0},clip,width=0.31\textwidth]{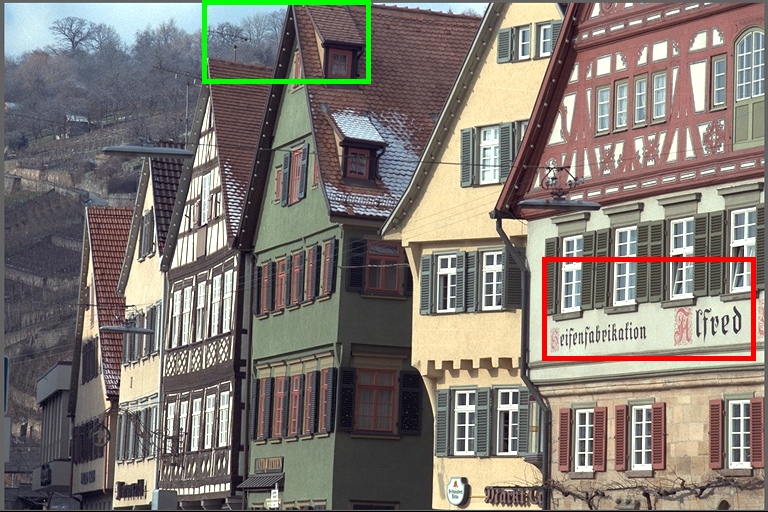}\\\texttt{kodim08.png}
	\end{tabular}
	\begin{tabular}{ccc}
		\begin{tikzpicture}
			\node[anchor=south west,inner sep=-0.1] (img) at (0,0) {\includegraphics[width=0.18\textwidth]{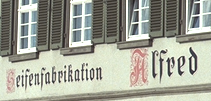}};
			\begin{scope}[x={(img.south east)},y={(img.north west)}]
				\draw[red] (0, 0) rectangle (1, 1); 
			\end{scope}
		\end{tikzpicture} &
		\begin{tikzpicture}
			\node[anchor=south west,inner sep=-0.1] (img) at (0,0) {\includegraphics[width=0.18\textwidth]{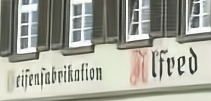}};
			\begin{scope}[x={(img.south east)},y={(img.north west)}]
				\draw[red] (0, 0) rectangle (1, 1); 
			\end{scope}
		\end{tikzpicture} & \hspace{-8pt}
		\begin{tikzpicture}
			\node[anchor=south west,inner sep=-0.1] (img) at (0,0) {\includegraphics[width=0.18\textwidth]{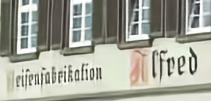}};
			\begin{scope}[x={(img.south east)},y={(img.north west)}]
				\draw[red] (0, 0) rectangle (1, 1); 
			\end{scope}
		\end{tikzpicture} \\
		\begin{tikzpicture}
			\node[anchor=south west,inner sep=-0.1] (img) at (0,0) {\includegraphics[width=0.18\textwidth]{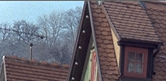}};
			\begin{scope}[x={(img.south east)},y={(img.north west)}]
				\draw[green] (0, 0) rectangle (1, 1); 
			\end{scope}
		\end{tikzpicture} &
		\begin{tikzpicture}
			\node[anchor=south west,inner sep=-0.1] (img) at (0,0) {\includegraphics[width=0.18\textwidth]{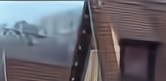}};
			\begin{scope}[x={(img.south east)},y={(img.north west)}]
				\draw[green] (0, 0) rectangle (1, 1); 
			\end{scope}
		\end{tikzpicture} & 
		\begin{tikzpicture}
			\node[anchor=south west,inner sep=-0.1] (img) at (0,0) {\includegraphics[width=0.18\textwidth]{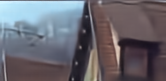}};
			\begin{scope}[x={(img.south east)},y={(img.north west)}]
				\draw[green] (0, 0) rectangle (1, 1); 
			\end{scope}
		\end{tikzpicture}\\
		PSNR / BPP & $25.3827/0.2424$ & $24.7371/0.2424$ 
	\end{tabular}
	\begin{tabular}{c}
		\includegraphics[width=0.31\textwidth,trim={0 2.5cm 0 1.5cm},clip]{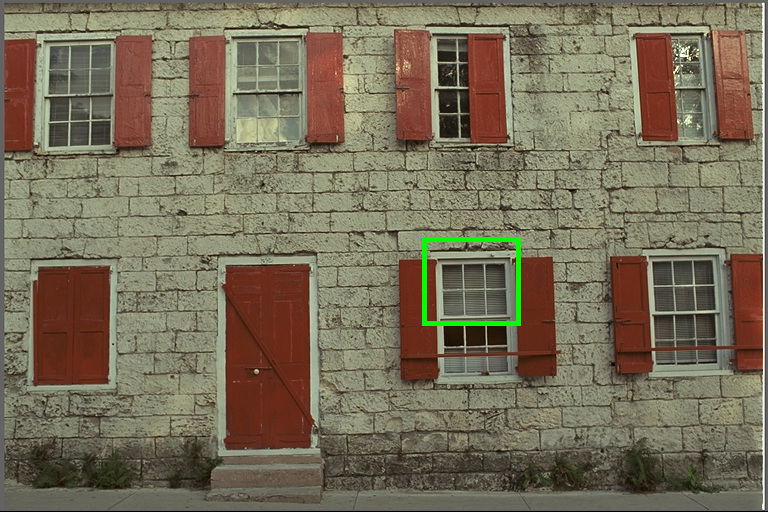}\\\texttt{kodim01.png}
	\end{tabular}
	\begin{tabular}{ccc}
		\includegraphics[width=0.18\textwidth]{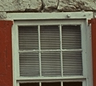} &
		\includegraphics[width=0.18\textwidth]{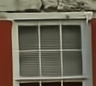} & 
		\includegraphics[width=0.18\textwidth]{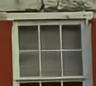} \\
		PSNR / BPP & $27.2582/0.2877$ & $27.0812/0.3135$ 
	\end{tabular}
	\begin{tabular}{c}
		\includegraphics[width=0.31\textwidth,trim={0 4.5cm 0 2cm},clip]{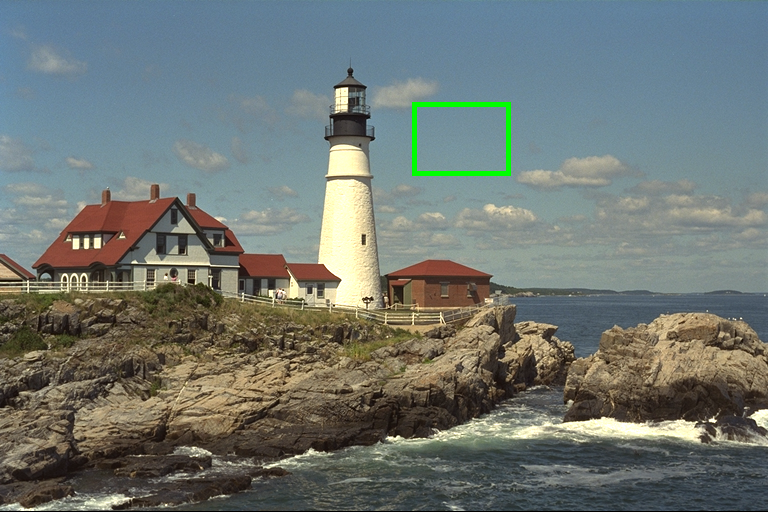}\\\texttt{kodim21.png}
	\end{tabular}
	\begin{tabular}{ccc}
		\includegraphics[width=0.18\textwidth]{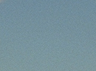} &
		\includegraphics[width=0.18\textwidth]{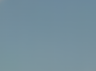} & 
		\includegraphics[width=0.18\textwidth]{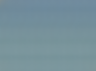} \\
		PSNR / BPP & $29.2820/0.2157$ & $28.9629/0.2159$ 
	\end{tabular}
	\caption{Visual evaluation of reconstructed \texttt{kodim08.png}, \texttt{kodim01.png} and \texttt{kodim21.png} images, respectively, in Kodak dataset. Crops are taken from ground truth, \network[] version of joint autoregressive network with GDN, and original joint autoregressive models, respectively. For \texttt{kodim08.png}, the crops are taken from the models trained with $\lambda=0.0018$, and the others are taken from the models with $\lambda=0.0035$ in the rate-distortion loss.}
	\label{fig:comp_visual_comp}
\end{figure*}

Comparison of our ELIC-\neuron[] model vs. original ELIC is depicted in Fig.~\ref{fig:comp_elic_rd_and_saving}. The RD curve for the anchor model Cheng et al.~\cite{cheng2020learned} is taken from the CompressAI benchmark \cite{begaint2020compressai}. As Fig.~\ref{subfig:comp_elic_rd} indicates, ELIC-\neuron[] surpasses the RD performance of ELIC. Remarkably, this is achieved by replacing classical convolution layers with \layer[] only in the image decoder and context model parts. More interestingly, ELIC-\neuron[] surpasses its ancestor even with significantly fewer layers compared to the~original model. This superiority can also be seen in Fig.~\ref{subfig:comp_elic_saving}. Even with the total number of layers reduced, ELIC-\neuron[] saves more than $1\%$ bit rate compared to the original ELIC. These results quantitatively show that the~superior representative power of \neuron[s] due to their strong inherent non-linearity and expanded receptive fields via feature shifting makes it possible to reduce the number of layers by one-third when convolutional layers are substituted by~\layer[s].

Visual examples of model outputs are given in Figure~\ref{fig:comp_visual_comp}. For~the crop shown by the red rectangle in \texttt{kodim08.png}, the result of MBT-\neuron[] is cleaner and sharper compared to the output of the original MBT-2018 model. Also, the lines of window shutters appear cleaner in the output of MBT-\neuron[], whereas in the original model's output, that region is smooth. In addition, the calligraphic A contains slightly more details in the output of the MBT-\neuron[] model. For the green crops in the same image, MBT-\neuron[] is able to reconstruct the horizontal parallel lines on the roof, which the original model fails to do so, and the top of the antenna pole, which is more faded in the output of the MBT-2018 model. The same difference in reconstruction power can be seen in the crops extracted from \texttt{kodim01.png} image, in which the output of MBT-\neuron[] has some clear parallel lines for the window blinds and the anchor model does not. MBT-\neuron[] also appears superior in areas with fewer details. In image \texttt{kodim21.png}, the crop taken from the sky has darker stripes and a minor color shift in the~output of off-the-shelf model whereas the fidelity is better preserved in the output of MBT-\neuron[].

\subsection{Image Classification}
\label{subsec:imgclass}

\subsubsection{Architecture}
\label{sssec:imgclass_arch}

For this traditional computer vision task, we have chosen the well-known ResNet20 architecture \cite{he2016deep}, having $20$ layers in total,  and show results on the CIFAR10 dataset~\cite{krizhevsky2009learning}. The network starts with a convolution layer having $16$ filters with $3\times3$ kernels, followed by batch normalization~\cite{ioffe2015batch}, which maintains the $32\times32$ spatial dimensions of  input images. The core architecture comprises of three stages of residual blocks with $[3, 3, 3]$ blocks per stage, progressively increasing channel dimensions from $16$ to $32$ to $64$ while reducing spatial resolution from $32\times32$ to $16\times16$ to $8\times8$ through strided convolutions. Each basic residual block contains two $3\times3$ convolutions with batch normalization, using a shortcut connection that adds the input to the block output. When dimensions mismatch, the shortcut employs a $1\times1$ convolution with batch normalization. \relu\ is applied after the~first convolution and after the residual addition, following the post-activation design. The output stage employs global average pooling and a fully connected layer for $10$-class prediction. All convolutional layers omit bias terms as they are followed by batch normalization. For future reference, we denote this architecture as ResNet(3,3,3).

We then introduce our PadéResNet, where all convolutional layers are replaced with  \layer[] layers, all \relu\ activations are removed, and the final fully connected layer is converted to a Padé linear layer, all using \neuron[]$^S_{[1/1]}$ neurons. In order to demonstrate layer efficiency, we reduced the number of residual blocks to 2, creating PadéResNet(2,2,2) without any \shifter module. This configuration means two residual blocks operate with $16$ channels, followed by two with $32$ channels, and another two with $64$ channels, totaling $14$ layers, with everything else staying the same as the original ResNet. To further demonstrate the performance, we incorporate the second \shifter version into every convolutional layer within each residual block. This model is referred as PadéResNet-II(2,2,2). Finally, we chose to further reduce the number of blocks introducing PadéResNet-II(1,1,2) which has a total of $10$ layers.

\subsubsection{Training Details}
\label{sssec:imgclass_train_Detail}

We train all models from scratch using $32\times32$ patch size, cropped from images that were padded by $4$, with batch size of $250$. The AdamW optimizer \cite{loshchilov2017decoupled} is employed with a learning rate of $10^{-3}$, a weight decay of $5\times10^{-4}$, and a cosine annealing scheduler for $600$ epochs, until the learning rate reached $2\times10^{-6}$. Our data augmentation strategy involved padding each image by $4$ on all sides, followed by taking random $32\times32$ crops. Additionally, we incorporated random horizontal and vertical flips, $90$-degree rotations, channel shuffling, and the introduction of $40$ dB SNR Gaussian noise. A validation set of $5,000$ images was randomly separated from the training set. The results are presented in Table \ref{tbl:imgclass_results_quant}.

\subsubsection{Comparison of Model Performance}
\label{sssec:imgclass_results}

\begin{table}
	\centering
	\caption{Accuracy results for different image classification models trained and tested on the CIFAR10 dataset.}
		\begin{tabular}{|c||c|}
			\hline
			Model & Accuracy \\\hhline{|=||=|}
			ResNet(3,3,3) & $84.70\%$ \\\hline 
			PadéResNet(2,2,2) & $85.07\%$ \\\hline 
            PadéResNet-II(2,2,2) & $85.97\%$ \\\hline 
            PadéResNet-II(1,1,2) & $84.93\%$ \\\hline 
		\end{tabular}
	\label{tbl:imgclass_results_quant}
\vspace{-6pt}
\end{table}

These results clearly show that a network with $10$ layers using \neuron[]$^S$ neurons and the second shifting strategy still surpasses the performance of the base model ResNet(3,3,3), which has a total of $20$ layers. The results support our claim that Paon-S neurons lead to layer-efficient architectures that require less amount of sequential operations, thus provide the possibility of faster inference.

To test the resilience of \neuron[s] to  lower precision implementation, we train PadéResNet-II(2,2,2) architecture in \texttt{float16} and \texttt{bfloat16} data types, which are used for faster training and reduced storage requirements. Surprisingly, the accuracy values of these models are computed as $86.30\%$ and $86.52\%$, respectively, indicating  \neuron[s] show strong performance even with lower precision training.

\section{Conclusion}
\label{sec:conc}

We propose a novel inherently non-linear neuron model called the Padé approximant neuron or in short \neuron[]. \neuron[s], supported by the well-known Padé approximation theory, possess stronger non-linear approximation capability with only a few layers compared to classical neurons, which need many layers to approximate a non-linear function by a cascade of piecewise linear functions. We further propose a smoothed variant called \neuron[]$^S$ to alleviate the potential singularity problem of rational function approximations in order to achieve a more continuous mapping. Interestingly, \neuron[]$^S$ achieves even stronger non-linearity than \neuron[] with the same number of parameters and similar complexity.

The main advantages of \neuron[]$^S$ can be summarized as: i) \neuron[]$^S$ provides strong non-linearity without a need for additional fixed non-linearity (e.g., ReLU, GeLU) or learned non-linearity (e.g., GDN). ii) \neuron[]$^S$ provides diversity of non-linearity as each \neuron[]$^S$ learns a different non-linearity. iii)~\neuron[]$^S$ provides layer-efficiency, i.e., has stronger non-linear approximation capacity with only a few layers.

Network layers constructed by \neuron[]$^S$ are called Padé Layers (\layer[]). We can construct convolutional \layer[] or fully-connected \layer[]. The receptive field of \neuron[]$^S$ in convolutional \layer[] can be increased using an approach similar to the well-known deformable convolutions as in the case of classical neurons. To this effect, we introduce two different \shifter methods: kernel-wise shift (as a group) and element-wise shift.

Experimental results provide strong evidence on the superiority of convolutional \layer[] over classical convolutional layers. Experiments on SISR, image compression and image classification quantitatively and qualitatively demonstrate that direct replacement of classic convolutional layers with \layer[s] improves the performance of benchmark models, such as ResNet, Minnen2018~\cite{minnen2018joint} and ELIC~\cite{he2022elic} with fewer number of layers compared to the original benchmark models. It is important to note that our compression model MBT-\neuron[] does not need GDN layers, which means significant savings on the parameter count and complexity.

\network[s] are beneficial in scenarios, where inference time is a primary concern.           The layer-efficiency of \network[s] makes them \textquotedblleft smaller\textquotedblright in terms of the number of sequential operations, which translates into shorter inference times with efficient GPU/TPU implementations. Furthermore, their resilience to lower precision implementations makes them suitable for possible real-world deployment.

Despite its important advantages, \neuron[s] also have some limitations. First, efficient implementation of \neuron[]$^S$ for real-world deployment demands expert coding. Optimizing polynomial division and associated operations for various platforms (e.g., GPUs, TPUs, mobile chips) requires careful low-level programming to maximize throughput and minimize latency. Second, training \network[s] with learned non-linearity can be slower compared to networks using simpler fixed activation functions. The non-linear nature of polynomial division can introduce optimization complexities, leading to slower convergence rates, which can be mitigated by clever hyperparameter selection.

The core advantage of \network[s] lies in their ability to achieve comparable or superior performance with a small number of layers compared to the classic neuron model. On the other hand, deeper networks with \relu\ activations might already achieve the necessary \textquotedblleft degree\textquotedblright  of non-linearity by approximating it via fine-granular piece-wise linear functions, which  makes the expressive advantage of \neuron[]$^S$ with learned nonlinearity less advantageous. Therefore, while \neuron[s] are promising to design layer-efficient networks, the competitive advantage of \neuron[s] can be less if one shall implement deep networks.

\bibliographystyle{IEEEbib}
\bibliography{refs}


%

\begin{IEEEbiography}[{\includegraphics[width=1.2in,height=1.5in,trim={20 30 18 0},clip,keepaspectratio]{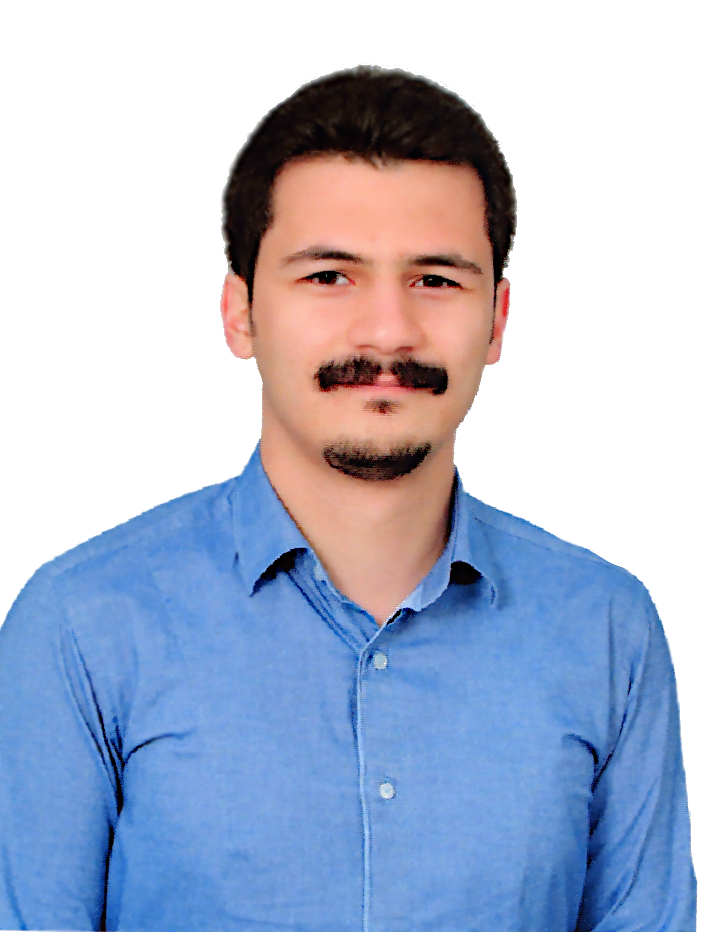}}]{Onur Keleş}
received B.Sc. and M.Sc. degrees in Electrical and Electronics Engineering from Boğaziçi University, İstanbul, Türkiye, in 2016 and 2019, respectively. He received his Ph.D. degree in Electrical and Electronics Engineering from Koç University, İstanbul, Türkiye, advised by A. Murat Tekalp, in 2025. He is currently with Codeway Digital Services, İstanbul, Türkiye, as Senior AI Research Scientist.
\end{IEEEbiography}
\begin{IEEEbiography}[{\includegraphics[width=1in,height=1.25in,clip,keepaspectratio]{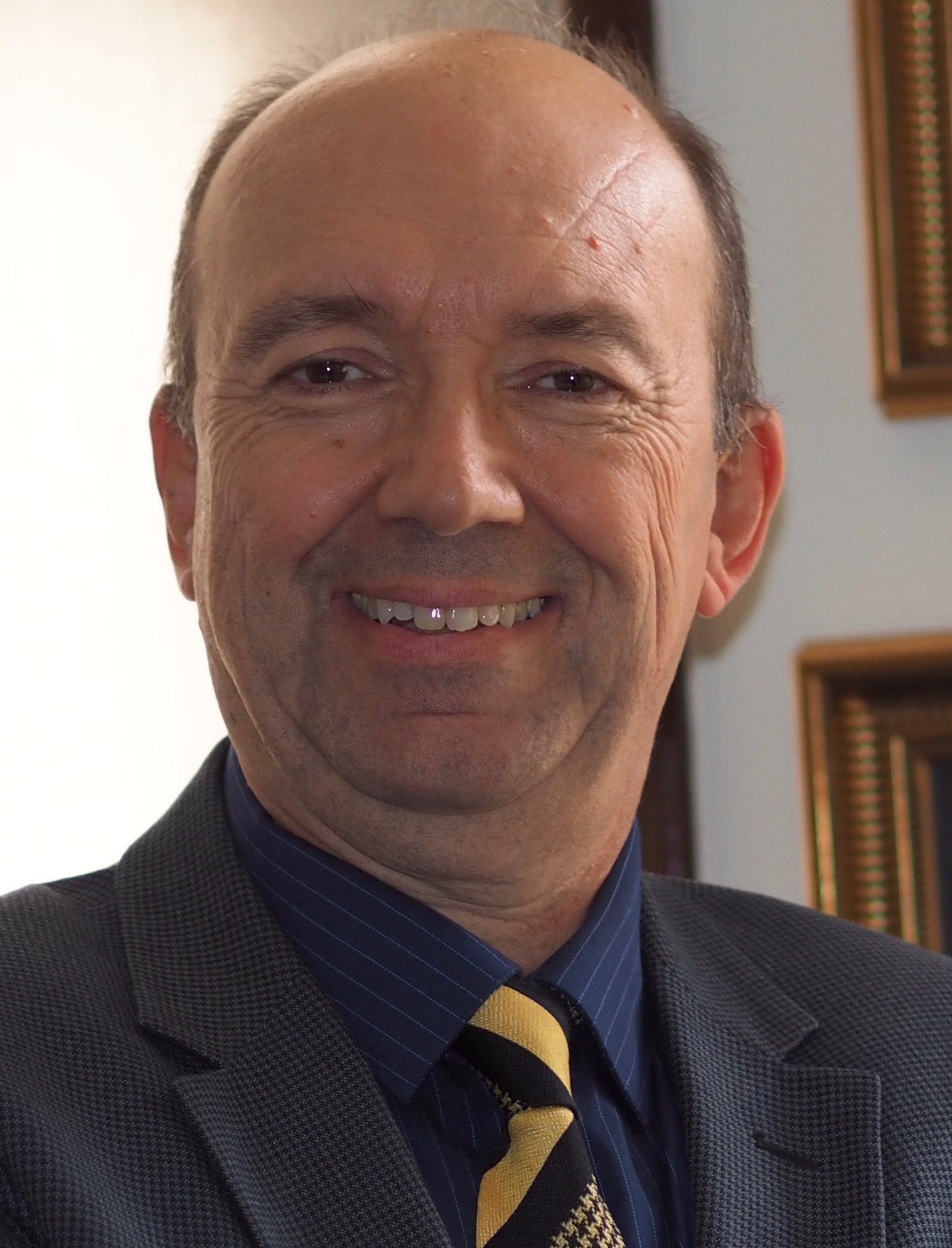}}]{A. Murat Tekalp} (S'80-M'84-SM'91-F'03) received Ph.D. degree in Electrical, Computer, and Systems Engineering from Rensselaer Polytechnic Institute (RPI), Troy, New York, in 1984, He was with Eastman Kodak Company, Rochester, New York, from 1984 to 1987, and with the University of Rochester, Rochester, New York, from 1987 to 2005, where he was promoted to Distinguished University Professor. He is currently Professor at Koc University, Istanbul, Turkey. He served as Dean of Engineering between 2010-2013. His research interests are in digital image and video processing, including video compression and streaming, video networking, and deep learning for image/video processing.
	
He has been elected a member of Turkish Academy of Sciences and Academia Europaea. He served as Associate Editor for IEEE Trans. on Signal Proc. (1990-1992) and IEEE Trans. on Image Proc. (1994-1996). He was the~Editor-in-Chief of the EURASIP journal Signal Proc.: Image Comm. published by Elsevier (1999-2010). He was on the Editorial Board of IEEE Signal Processing Magazine (2007-2010), Proceedings of the~IEEE (2014-2020), and Wiley-IEEE Press (2018-2024). He chaired IEEE Signal Processing Society Technical Committee on Image and Multidim. Signal Processing (Jan. 1996 - Dec. 1997). He was appointed as the General Chair of IEEE Int. Conf. on Image Processing (ICIP) in 2002, and as the Technical Program Co-Chair for IEEE ICIP~2020 and ICIP~2024. He is serving in the European Research Council (ERC) Panels since 2009. Dr. Tekalp authored the Prentice Hall book Digital Video Processing~(1995), second edition~(2015). 
\end{IEEEbiography}




\end{document}